\documentclass[journal,twoside,web]{ieeecolor}
\usepackage{etoolbox}
\makeatletter
\@ifundefined{color@begingroup}%
{\let\color@begingroup\relax
\let\color@endgroup\relax}{}%
\def\fix@ieeecolor@hbox#1{%
\hbox{\color@begingroup#1\color@endgroup}}
\patchcmd\@makecaption{\hbox}{\fix@ieeecolor@hbox}{}{\FAILED}
\patchcmd\@makecaption{\hbox}{\fix@ieeecolor@hbox}{}{\FAILED}
\usepackage{jsen}
\usepackage{cite}
\usepackage{amsmath,amssymb,amsfonts}
\usepackage{algorithmic}
\usepackage{graphicx}
\usepackage{textcomp}
\usepackage{wrapfig}
\usepackage{hyperref}
\usepackage{textcomp}
\usepackage{blindtext}
\usepackage{multirow}
\usepackage{makecell}
\usepackage{soul}

\def\BibTeX{{\rm B\kern-.05em{\sc i\kern-.025em b}\kern-.08em
    T\kern-.1667em\lower.7ex\hbox{E}\kern-.125emX}}
\definecolor{abstractbg}{rgb}{0.89804,0.94510,0.83137}
\begin{document}
\title{TacPalm: A Soft Gripper with a Biomimetic Optical Tactile Palm for Stable Precise Grasping}
\author{Xuyang Zhang, Tianqi Yang, Dandan Zhang, Nathan F. Lepora
\thanks{The authors acknowledge the financial support of the ISCF Research Center for Smart, Collaborative Industrial Robotics (EP/V062158/1).}
\thanks{XZ is with the Department of Engineering, King's College London. TY is with Visual Information Laboratory, University of Bristol. D. Zhang is with the Department of Bioengineering, Imperial College London. NL is with the Department of Engineering Mathematics and Bristol Robotics Laboratory, University of Bristol, U.K.}
\thanks{Email: k21178779@kcl.ac.uk, tianqi.yang@bristol.ac.uk, d.zhang17@imperial.ac.uk, n.lepora@bristol.ac.uk}}

\IEEEtitleabstractindextext{%
\fcolorbox{abstractbg}{abstractbg}{%
\begin{minipage}{\textwidth}%
\begin{wrapfigure}[15]{r}{3.4in}%
\includegraphics[width=3.3in]{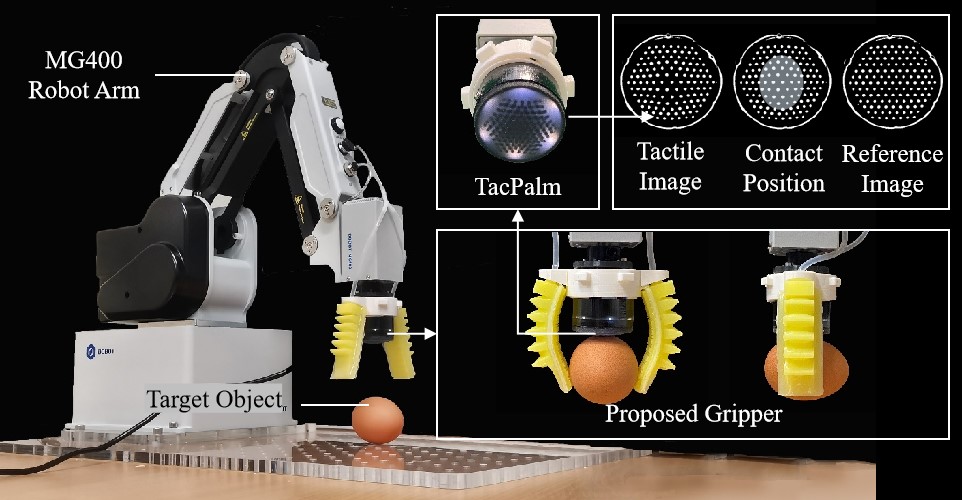}%
\end{wrapfigure}%
\begin{abstract}
Manipulating fragile objects in environments such as homes and factories requires stable and gentle grasping along with precise and safe placement. Compared to traditional rigid grippers, the use of soft grippers reduces the control complexity and the risk of damaging objects. However, it is challenging to integrate camera-based optical tactile sensing into a soft gripper without compromising the flexibility and adaptability of the fingers, while also ensuring that the precision of tactile perception remains unaffected by passive deformations of the soft structure during object contact. In this paper, we demonstrate a modular soft two-fingered gripper with a 3D-printed optical tactile sensor (the TacTip) integrated in the palm. We propose a soft-grasping strategy that includes three functions: light contact detection, grasp pose adjustment and loss-of-contact detection, so that objects of different shapes and sizes can be grasped stably and placed precisely, which we test with both artificial and household objects. By sequentially implementing these three functions, the grasp success rate progressively improves from 45\% without any functions, to 59\% with light contact detection, 90\% with grasp pose adjustment, and 97\% with loss-of-contact detection, achieving a sub-millimeter placement precision. Overall, this work demonstrates the feasibility and utility of integrating optical tactile sensors into the palm of a soft gripper, of applicability to various types of soft manipulators. The proposed grasping strategy has potential applications in areas such as fragile product processing and home assistance.
\end{abstract}

\begin{IEEEkeywords}
soft gripper, tactile sensing, robot grasping
\end{IEEEkeywords}
\end{minipage}}}

\maketitle
\section{Introduction}
\label{sec:introduction}
\IEEEPARstart{I}{tems} in agriculture, manufacturing and home assistance vary greatly in shape, size and hardness \cite{chua2003robotic,erzincanli1997meeting}. The grasping, moving and placing of soft, delicate and fragile objects requires good adaptability, safety, high sensitivity, robustness and accuracy of the gripper \cite{lee2000tactile}. Traditional rigid two-finger grippers face challenges when seeking high compliance and adaptability without compromising grasping precision. In contrast, soft grippers' adaptability and passive compliance can enable safe, robust and reliable grasping of flexible and fragile items with a wide range of object properties~\cite{shintake2018soft,hughes2016soft}.


Meanwhile, the benefits of a tactile-sensorized gripper are becoming apparent, by adding important functionality to item handling such as reacting to object slippage~\cite{9252140,8460495} and direct feedback about the contact to aid grasping~\cite{wall2017method,8593528,james2021tactile, he2023tacmms}. Camera-based optical tactile sensors, such as the GelSight~\cite{yuan2017gelsight}, TacTip~\cite{ward2018tactip,lepora2021soft} and others~\cite{gomes2020geltip,jiang2023, zhang2024compact, zhang2022hardware, zhang2023novel, jiang2024rotipbot, zhangrotip}, offer a granularity of contact detail that is currently not possible to achieve with non-vision-based soft tactile technologies. For example, this granularity in tactile image detail is key to accurately estimating the pose of a contact feature (e.g. an edge or surface), which then enables robust tactile servoing or pushing manipulation of unknown objects~\cite{lepora2021pose,lloyd2021goal}. 

However, for soft grippers, it remains an open challenge to integrate such camera-based optical tactile sensors with soft fingers that can actively deform~\cite{deng2022design,he2020soft}. The main issue is that these sensors rely on internal camera modules that are rigid components with lighting assemblies and wiring, and so are incompatible with maintaining the adaptability and other benefits of soft grippers. That said, a couple of recent studies have successfully integrated optical tactile sensors into the fingertips of semi-flexible grippers\cite{she2020exoskeleton, liu2022gelsight}. Even so, it remains an open problem how to integrate camera-based tactile sensors into soft adaptable gripping technologies such as pneumatically-driven soft fingers.


The perspective introduced and explored in this paper is that camera-based optical tactile sensing can be useful when placed in part of the gripper that does not impede the adaptability and function of the soft components. In this we take inspiration from the soft palm of the human hand, which is adaptive and passively compliant and can help form a force-closure grasp on objects with closed fingers. Tactile perception at the palm provides information about the pose of a held object, which is useful information for a stable grasp. Researchers rarely consider integrating tactile perception at the palm when designing grippers, with the palm only acting as a base for connecting fingers \cite{townsend2000barretthand} or as a support surface for force grasping \cite{odhner2014compliant}. Some studies have implemented tactile perception on the palm with distributed pneumatic tactile sensors or force-sensor arrays \cite{9341691,8793842}, but the feedback of such tactile information is more limited than that provided by optical tactile sensing.

Integrating a camera-based tactile palm on a gripper with soft fingers raises the question of what benefits will tactile sensing bring to soft grippers. Here we propose that additional functionality stems from determining the initial contact on an object, in particular an accurate estimation of the contact pose. This pose estimate can then be used to guide the gripper to explore or move to a good location on the object for the soft fingers to grasp. We view this as analogous to how pose estimation from soft biomimetic optical tactile sensing enables accurate control of a robot for servoing or manipulation~\cite{lepora2021pose,lloyd2021goal}. A key aspect of this added functionality is that the soft fingers must be retractable away from the object so that only the tactile palm contacts initially, which motivates our use of soft pneumatic fingers designed to have that capability.

The main contributions of this paper are:\\
\noindent 1) We adapt a soft biomimetic optical tactile sensor (the TacTip~\cite{ward2018tactip,lepora2021soft}) into a tactile palm and rigid base to attach soft pneumatic fingers. These fingers are designed to have a workspace that can hold objects securely against the soft palm and retract to not impede the palm during initial contact. For this work, we consider a 2-fingered gripper, but the approach should extend to other morphologies, shown in Appendix II.\\
\noindent 2) For perceiving the location of the object from contact with the palm, we propose a tactile-based object-feature pose estimation model using a convolutional neural network on tactile images, with which we achieve a sub-millimeter position accuracy and degree-level angle accuracy.\\ 
\noindent 3) We introduce pre-grasp control based on tactile servoing to adjust the pose of the gripper relative to the object, applicable to curved- and flat-topped objects. We then show this pre-grasp adjustment improves the grasping performance and placement precision of the soft gripper for a range of household objects. \\
\noindent 4) We also consider the use of the tactile palm for contact detection and loss-of-contact detection, implemented with image structural similarity methods, which we show that adjusting the grasp force can further improve the performance.

All materials and methods will be openly released, including designs and instructions to fabricate the soft tactile palm/fingers and the software to operate the robot.

\section{Background}

Robot grasping and manipulation require particular grasping strategies and online adaptive adjustment according to the shape of the object. It is difficult for a gripper to perform stable grasping and precise placement without sensory capabilities, such as tactile feedback, or access to positional and physical information about the object \cite{teeple2020multi}. 

\subsection{Tactile sensing in soft grippers}

At present, several studies have integrated camera-based optical tactile sensors into the fingers of rigid hands, enabling the classification of unknown objects, prediction of the grasping success rate and slip detection \cite{8593528, james2021tactile, ward2016tactile, ward2017model, gomes2022geltip}. In contrast, research on tactile soft grippers has made little progress because it is difficult to integrate the rigid components such as camera modules, circuit boards and lighting into soft fingers while not impeding their active deformation.

Soft stretchable tactile sensors are being integrated into soft grippers, either along the neutral axis of the finger to measure its curvature changes, or on the surface of the soft finger to estimate the pressure or contact deformation \cite{deng2022design,hao2020soft, park2020sensorized,zuo2021soft}. The classification of the object can be achieved by analyzing the time series of tactile signals. However, this tactile feedback has low spatial acuity and so cannot accurately determine the position and posture of the contact to the precision of camera-based optical tactile sensing. To obtain richer contact information, flexible tactile array sensors with higher accuracy and spatial acuity have been integrated on the soft finger surface, to achieve a more stable grasp by controlling the pressure distribution on the contact surface \cite{li2021robot}. However, such an approach still does not reach the information content of vision-based tactile sensors, and moreover such technologies are not as readily fabricated as those based on camera technology. 

\subsection{Tactile palms in robot hands}
In grasping tasks, the palm of the hand is usually considered as a passive support surface \cite{9869694}. The compliance of the palm can be enhanced by using flexible materials (rubber, silicone), which also improves the grasping stability and safety. However, most grasps with the help of the palm are open-loop designs due to the lack of perception of the contact state, especially for soft grippers. 

Some studies have begun to integrate tactile sensing within the palm component of a gripper to estimate the contact state with the object during grasping. Li et al. affixed a flexible fabric pressure sensor to the palm for closed-loop control of the finger force by estimating the contact force at the palm \cite{9872269}. Shorthose et al. designed a distributed pneumatic tactile sensor array to classify object properties based on tactile feedback from fingers and palms, and adaptively adjusted the hand posture \cite{9706272}. Wang et al. designed a force-sensing array on the palm and detected object sliding by observing force signal changes \cite{8793842}. Lei et al. integrated a soft tactile sensor array in the palm to estimate the contact position and contact force, which improved the spatial acuity of tactile sensing to an average position estimation error of 2.3 mm \cite{9869694}. To achieve grasping stability and manipulation precision, the palm also needs to be highly compliant and have high-performance tactile perception capabilities. 

\subsection{Camera-based optical tactile sensing}

Several recent studies have integrated camera-based optical tactile sensors into soft grippers due to their high sensitivity, high information content and good synergies with deep learning methods. Some of these studies focus on integrating optical tactile sensors on semi-flexible fingers. Liu et al. integrated the GelSight tactile sensor on the fingertip of a Fin-Ray-inspired finger to perceive the object pose, and completed the task of grasping and then placing a glass \cite{liu2022gelsight}. They also proposed a novel three-finger robot hand that has optical tactile sensing along the entire length of each finger. The fingers are compliant, and constructed with a soft shell supported with a flexible endoskeleton \cite{liu2023gelsight}. She et al. designed an exoskeleton-covered tendon-driven soft gripper with a camera embedded at the finger joint to obtain tactile information on the surface, then achieved the classification of the size and shape of the grasped object \cite{she2020exoskeleton}. Lu et al. proposed a reconfigurable pneumatic gripper with a modular design and integrated the DigiTac tactile sensor into the fingertips, enabling the closed-loop control of the adaptable gripping \cite{lu2024dexitac}.

There are also some studies that design the optical tactile sensor itself as a soft gripper. TacEA proposed by Xiang et al. integrates a stretchable EA pad on an inflatable TacTip optical tactile sensor, achieving grasping by electro adhesion and classification of two-dimensional objects \cite{xiang2019soft}. Sakuma et al. designed a granular-jamming-based universal soft gripper, filled with transparent objects inside, embedded with a built-in camera to track the marks on the gripper surface, realizing 3D reconstruction of the contact surface \cite{sakuma2018universal}. The soft gripper TaTa proposed by Li et al. embeds RGB tricolor LEDs and uses a built-in camera to observe color changes on the contact surface to obtain tactile information, which significantly improves the information content of the tactile image \cite{li2022tata}. 

There are a few studies that integrated camera-based optical tactile sensors into the soft palm. Liu et al. developed a novel structurally compliant soft palm with tactile feedback, which enabled more surface area contact for the objects that are pressed into it \cite{liu2024passively}. Hu et al. propose a novel curved visuotactile sensor, the GelStereo Palm, serving as a soft palm, which senses the 3-D contact geometry on a curved surface using a binocular vision system, achieving accuracy and robustness of the 3-D contact geometry sensing \cite{hu2023gelstereo}. Zhang et al. proposed a vision-based tactile sensor embedded into the palm of the soft hand, enabling the soft hand to perceive object texture and temperature \cite{zhang2024palmtac}.

Therefore, the mainstream of the integration location is still on soft fingers, with less work focusing on integrating tactile sensors into the palms. Moreover, the functions enabled by tactile palms are mainly limited to object classification and texture perception. Object pose perception and grasping adjustment strategies based on pose estimation have not been investigated.

Here we combine a soft biomimetic optical tactile palm with pneumatic soft fingers using an optical tactile sensor, the BRL TacTip. The TacTip is a biomimetic marker-based tactile sensor that mimics the structure and function of human tactile skin \cite{lepora2021soft}. We also propose a soft grasping strategy based on tactile perception of object pose, which improves the capabilities of the soft gripper for stable grasping and precise placement of unknown objects. Notably, this work primarily explores tactile soft grasping strategies for rigid, fragile objects.

\section{Hardware Methods}

\begin{figure*}[t!]
	\centering
	\begin{tabular}[b]{@{}c@{}}	\includegraphics[width=2\columnwidth]{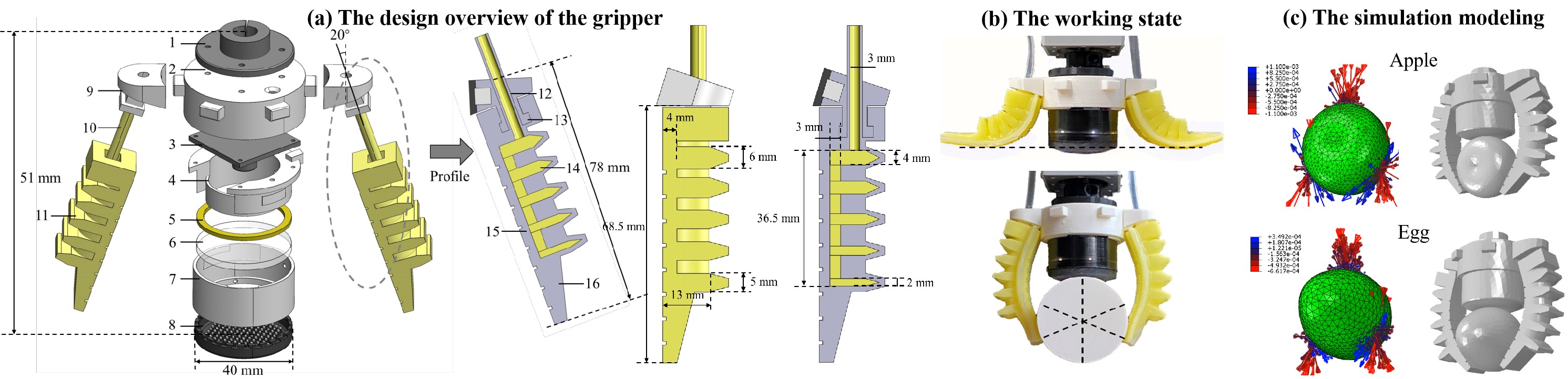} \\
	\end{tabular}
	\caption{ (a) Overview of the structural design. (1-Mounting flange. 2-Gripper body and housing. 3-Camera. 4-Camera mount. 5-LED ring. 6-Acrylic window. 7-Tip mount. 8-Flat skin with pins (markers). 9-Finger base. 10-Air tube. 11-Soft finger. 12-Air passage. 13-Embedded slot. 14-Air cavity. 15-Connected air passage. 16-Fingertip.) (b) The working state of the proposed soft tactile gripper. (c) Contact modelling of the proposed gripper. (Stress analysis and grasping deformation when grasping an apple (up) and an egg (down).}
	\label{fig:1}
	\vspace{-1em}
\end{figure*}

\subsection{Overall Design of the Soft Gripper with TacPalm}

Fig. \ref{fig:1} shows the structural design and working state of the proposed soft tactile gripper. The gripper consists of a mounting base, two parallel soft fingers and an optical tactile sensor, the BRL TacTip, mounted at the centre of the palm. It can be flange-mounted on different types of robotic arms to act as an end-effector. The modular design eases fabrication and maintenance, while allowing easy replacement of the tactile sensor and soft fingers to explore their capabilities. The tactile sensor is integrated into the mounting base by bolts and the soft fingers are connected to the base by slots. The fingers are driven by an air pump (Dobot MG400 MVPB Mini Vacuum Pump Box) that can achieve abduction and adduction movement, which is important for both bending the fingers onto and away from the palm. In the abduction state, the height of the inner surface of the finger is flush with the tactile skin, so that the finger does not interfere with the perception of objects. In the adduction state, the soft fingers compliantly grasp the object and press it firmly against the tactile skin. The soft fingers were fabricated using mould casting, while other parts of the gripper were fabricated by 3D printing. It is noteworthy that we chose the simplest parallel gripper configuration to validate the proposed tactile palm-based soft grasping strategy because we did not want to assume any prior information about the object and use it to design the soft gripper in a targeted manner.

\subsection{Base Design and Robot Mount}
The base of the TacTip normally just houses the camera and attaches by a mount onto a robot arm. For the tactile gripper, the base is customized to have rectangular projections spaced 60 degrees apart around the side of the cylindrical surface for connection with the soft fingers. The slotted connection facilitates finger replacement, and allows for changing different types and numbers of fingers, as well as the layout of the finger positions according to the task requirements.

The movement of the soft gripper is controlled by a desktop robot arm (Dobot MG400), which can realize translation in three dimensions and rotation around the vertical axis. The gripper is attached to the arm by a mounting flange that is standard for this arm. 

\subsection{Finger Design}
The structure of the finger is shown in the profile view in Fig. \ref{fig:1} (a), which is adapted from \cite{xavier2022soft}, and its overall size is similar to an adult human finger. Each finger consists of four parts: the main body, the cavity, the finger base, and the silicone air tube. The finger base is made by 3D printing, the main body is cast from Silicone RPRO 30 material (Silicone Rubber, made by Reschimica) through 3D printed moulds. We chose Shore A 30 material to achieve a balance between compliance and structural stability, which facilitates easier pneumatic control. The Shore A 20 silicone material, softer with good stretchability, led to local expansion and air leakage from the chambers under high pneumatic pressure. Conversely, the Shore A 40 silicone, harder with higher compressive strength, demanded increased pump pressure but lacked compliance, thereby hindering adaptation to object shapes. In power grasping scenarios, where the gripper lifts or moves objects, having softer fingers with inadequate grasping force can lead to changes in object posture due to gravity and inertia forces. Using a firmer material allows the fingers to withstand higher air pressure, enhancing their gripping power for heavier objects. The gently contoured fingertips offer ample surface contact and a firm grip, combined with the pliant tactile palm to ensure a stable grasp.

The fingertip is wedge-shaped to provide squeezing force and ensure enough contact area with the object. The inner surface of the finger is textured to minimize slip. The volume of the chambers decreases from the finger root to the fingertip, with larger chambers near the root to generate sufficient grasping force and smaller chambers near the tip to enhance contact compliance. The first chamber at the finger root is connected to a silicone air tube and all chambers are connected by channels inside the finger. The finger base is used for the connection with the mounting base, which is embedded into the finger root through a slot that becomes larger at the end to prevent the finger from slipping off. The silicone air tube and the finger base are cast together with the main body to ensure the connection is sealed for the pneumatics. 

\begin{figure}[t!]
	\centering
	\begin{tabular}[b]{@{}c@{}}
        \includegraphics[width=1\columnwidth,trim={0 0 0 0},clip]{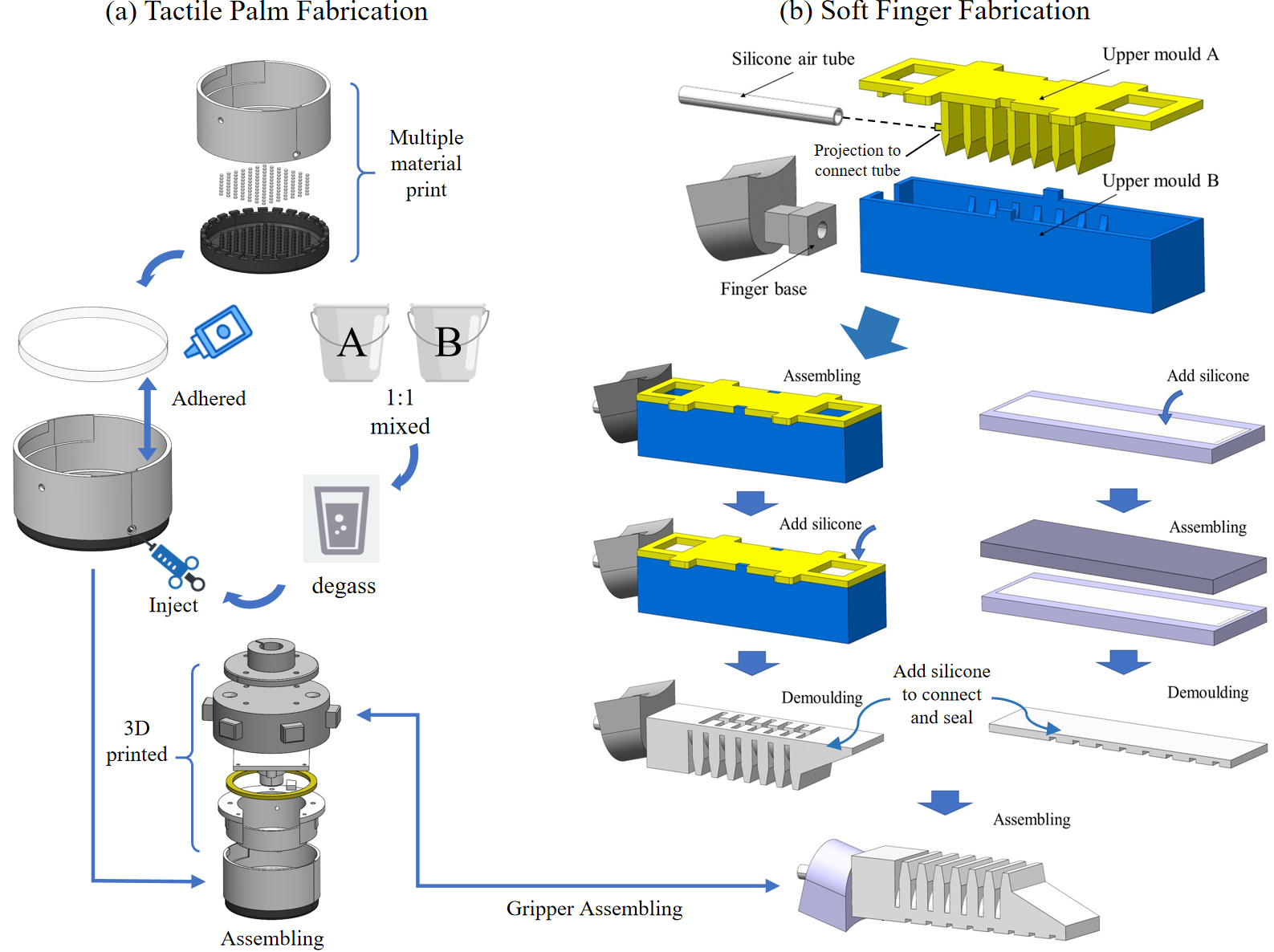} 
    \end{tabular}
	\caption{\textcolor{black}{The fabrication process of the soft gripper.}}
	\label{fig:2}
	\vspace{-1em}
\end{figure} 

\subsection{Tactile Sensor Design}
The sensor design is based on a low-cost soft biomimetic optical tactile sensor, the BRL TacTip~\cite{ward2018tactip,lepora2021towards}. In human skin, the outer epidermis and underlying papillary dermis form a network that transmits the skin deformation to nerve endings, creating tactile sensations. The human palm offers soft, stable support for grasped objects, ensures a secure grip in coordination with the fingers, with tactile feedback essential for sensing posture and shape. The proposed tactile palm mimics this by 3D printing a soft outer shell as the epidermis and filling the interior with elastomer gel as the dermis. Uniformly distributed pins replicate the papillary micro-structures, with an internal camera tracking the movement of these pins to transmit tactile signals. The proposed soft grasping strategy mimics human palm functions by utilizing the palm's tactile-based posture estimation capability. To be suitable as a palm, we designed the tactile skin with a curvature of 41.5\, mm, which is nearly flat but retains a slight curvature. This design allows for a maximum indentation of 5 mm, enabling effective tactile deformation when in contact with both curved surfaces and edges. Considering the size range of the grasped object and the spacing between the parallel fingers, we set the diameter of the tactile skin to 40 mm. The sensor features 127 white pins arranged in a hexagonal pattern, evenly distributed across the surface. A 170-degree fisheye lens was chosen for the camera module (ELP 2 Megapixel USB Camera), which has a resolution of 1920$\times$1080 (Full HD) and a frame rate of 30\, fps. We utilized an LED ring with six white LEDs evenly distributed to provide uniform internal illumination. The sensor's skin is made of a rubber-like material (Agilus30™; Shore A hardness 33), chosen for its excellent flexibility, wear resistance, and opacity. We employed TechsiL RTA27905 A/B to fill the sensor, which is a transparent elastic gel that is exceptionally soft (~Shore OO 10), incompressible, and leak-resistant, ensuring that the outer skin quickly returns to its original shape after deformation.

\subsection{Fabrication}
As shown in Fig. \ref{fig:2} (a),  we utilized a multi-material 3D printer (Stratasys J826 Prime) to simultaneously fabricate the sensor's outer skin, pins, and tip mount, ensuring seamless integration. A transparent, laser-cut acrylic window was adhered to the tip mount to create a cavity for gel injection. The transparent gel (Mix with a 1:1 ratio and degas) was then manually injected into the cavity through a filling port on the mount wall using a syringe. After the injection, the filling port was sealed with a plug. Once the gel had cured, the sensing components were assembled with the light source and camera module.

The manufacturing process of the finger is illustrated in Fig. \ref{fig:2} (b). The finger comprises a main body layer and a base layer, which are cast using separate moulds and subsequently bonded together using silicone. All casting moulds were produced using a 3D printer (Ultimaker 2) and coated with a release agent (Ease Release™ 200, Smooth-On, Inc.) to facilitate demolding. We first assemble the mould to cast the finger: A silicone tube passes through the finger base and connects to a protrusion on the upper mould A, allowing it to connect with an air chamber formed by upper mould A after demolding. The R PRO 30 silicone is then mixed in a 1:1 ratio, degassed using a vacuum pump, and poured into the assembled mould. The silicone is allowed to cure at room temperature for 3 hours. After demolding the finger components, we apply the same silicone to the connection area between the upper and lower parts of the soft finger to connect both and ensure a seal, and then assemble these parts.

Finally, we assembled all the components, including two soft fingers and the tactile palm. The fingers were connected to an air pump via pneumatic tubes, while the LED ring and camera were connected to the laptop via USB. The entire soft gripper was mounted onto the robotic arm through the mounting flange.

\subsection{Simulation Modeling} When exploring the gripper design, we modelled the gripper by finite element simulation method using Abaqus/CAE. The material parameters used for the finger based on the Yeoh model was set to \textit{C10=0.11}, \textit{C20=0.02}, \textit{C30=0.02}, while the soft palm was based on the Neo-Hookean hyperelastic model, with \textit{C10=0.0335}, \textit{D1=1.2297}. The finger root was fixed and a 15 kPa pressure load was applied to the inner wall of the chamber, while gravity was also applied to recreate the physical working environment. As shown in Fig. \ref{fig:1} (c), it can be observed from the deformation results and stress distribution that the three-directional surface support provided by the gripper forms a stable enclosed grasp, while the soft tactile palm provides sufficient adaptability.

\begin{figure}[t!]
	\centering
	\begin{tabular}[b]{@{}c@{}}
        \includegraphics[width=0.9\columnwidth,trim={0 0 0 0},clip]{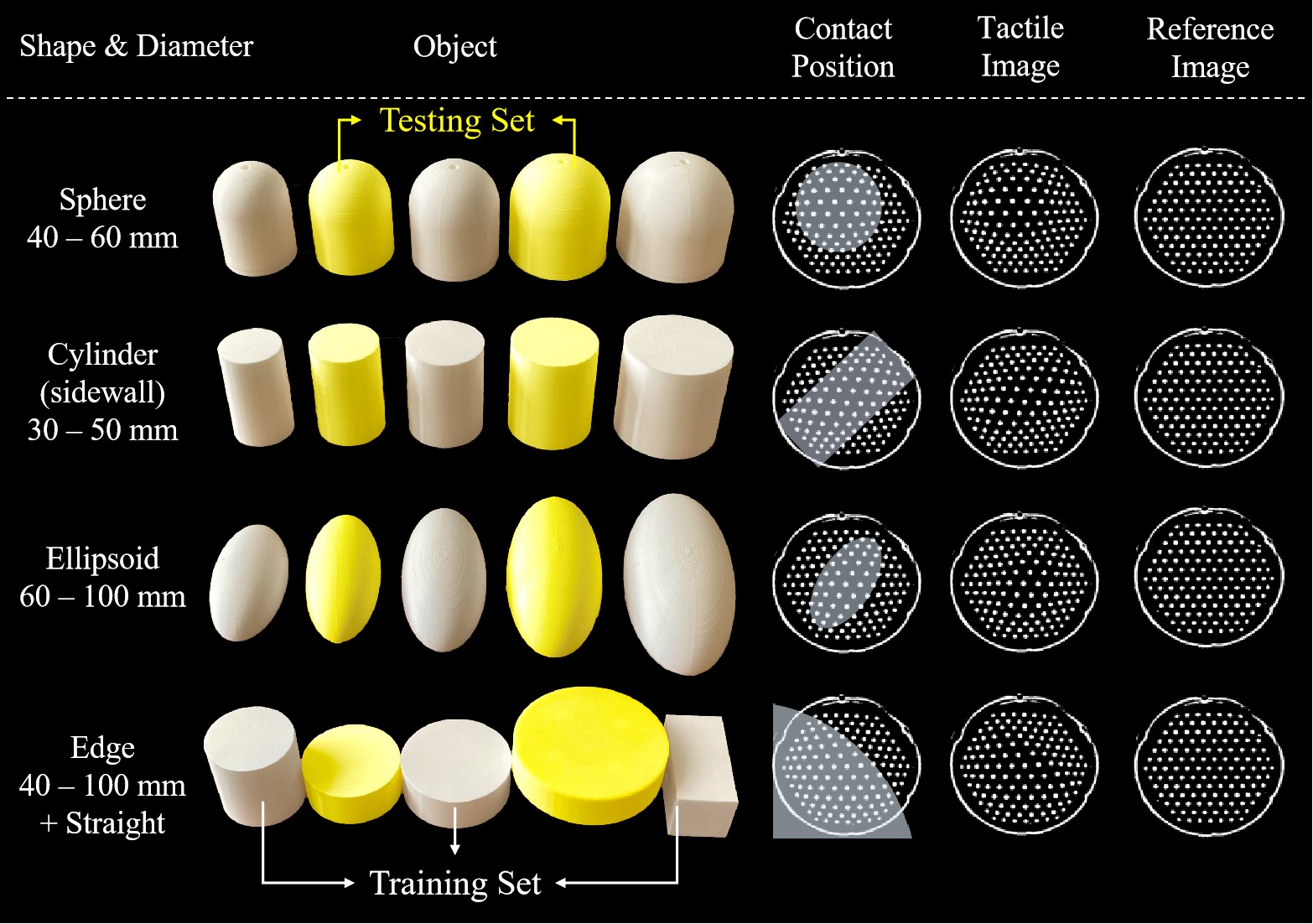} 
    \end{tabular}
	\caption{\textcolor{black}{3D-printed object set used for training and testing.}}
	\label{fig:3}
	\vspace{-1em}
\end{figure} 

\subsection{3D-printed Object Set}
As shown in Fig. \ref{fig:3}, objects of different shapes and sizes were 3D printed for data collection and testing, and their upper surfaces are of two categories based on their geometric features: curved surfaces (spheres, ellipsoids, and the sidewalls of cylinders) and edged surfaces (straight and curved boundaries). For each shape, we printed five objects of different sizes, matching the size differences between categories, three of which (white objects in Fig. \ref{fig:3}) were used to collect the dataset for training and the other two (yellow objects in Fig. \ref{fig:3}) to test the generalization ability of the trained model. The size range of printed models covers the dimensions of commonly used household objects graspable with one hand. Here we consider treating straight edges as a special case of curved edges for our pose estimation model. Including straight edges in the training dataset ensures that the trained model covers a wider range of curvature features for edges, thereby enhancing the model's generalization capability. Therefore, cuboids have been added to the training dataset.

\section{Algorithmic Methods}

\subsection{Overall Grasping Strategy}
\begin{figure}[t!]
	\centering
	\begin{tabular}[b]{@{}c@{}}
        \includegraphics[width=0.9\columnwidth,trim={0 0 0 0},clip]{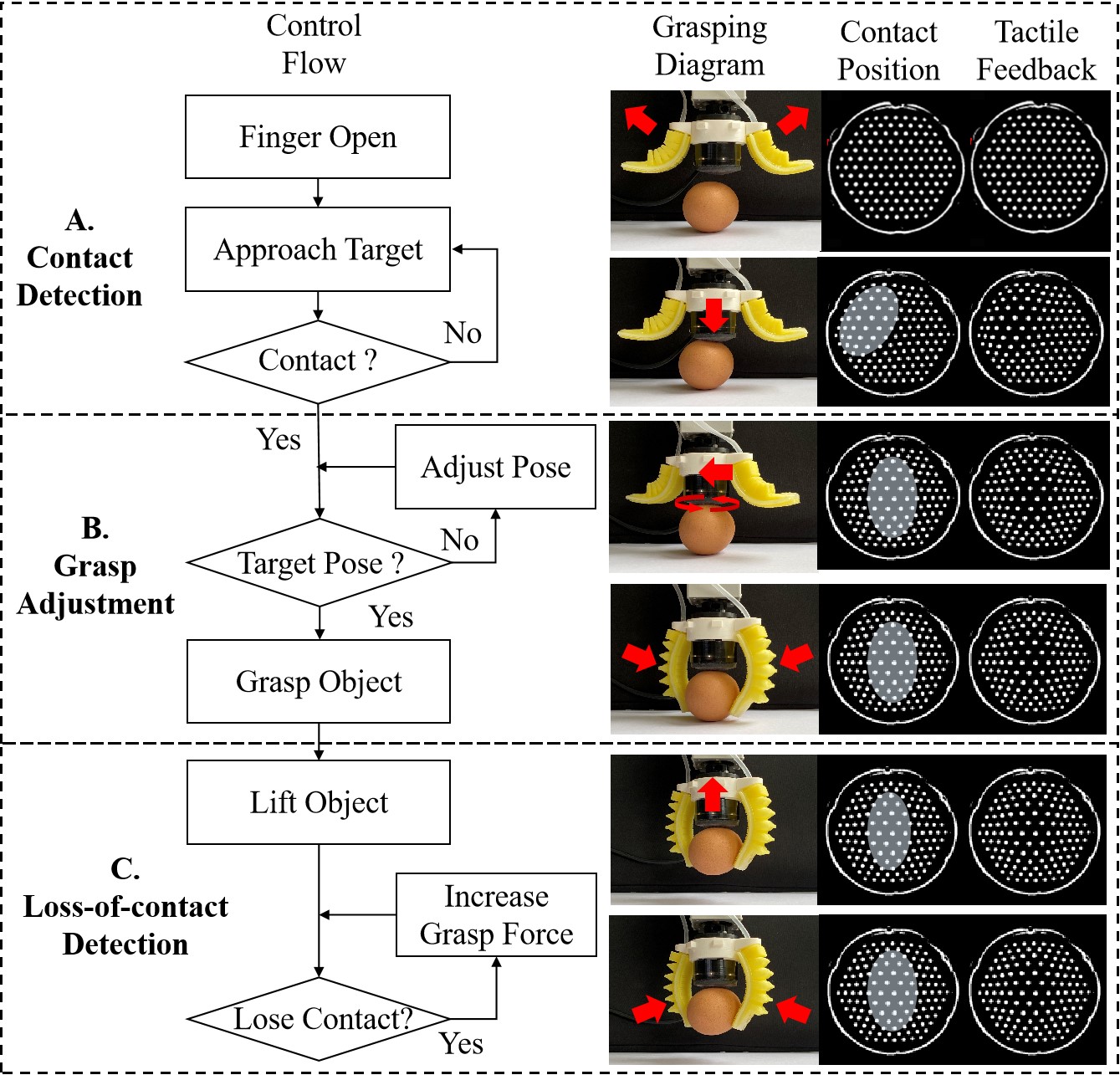} 
    \end{tabular}
	\caption{\textcolor{black}{Control flow of the soft grasping strategy, which includes the light contact detection, grasp pose adjustment and loss-of-contact detection, ensuring a stable and robust object grasping.}}
	\label{fig:4}
	\vspace{-1em}
\end{figure}

The flow chart of the proposed soft grasping strategy is shown in Fig. \ref{fig:4}. The functions achieved by the tactile sensing from the palm include: light contact detection, grasp pose adjustment and loss-of-contact detection. The soft gripper executes the same control process when grasping different objects, but implementing different pose adjustment strategies in the grasp adjustment stage according to the geometric features of the object. In particular, both the initial light-contact detection and later loss-of-contact detection are implemented using the Structural Similarity Index Measure (SSIM)~\cite{wang2004image}, an established measure from computer vision. Grasp pose adjustment is implemented by a CNN deep neural network that predicts object pose from a tactile image.

\subsection{Light Contact Detection}
Previous studies have demonstrated that the SSIM is a simple and robust method for measuring differences between two tactile images \cite{james2021tactile, lepora2021towards}. The output of the SSIM is a real number within [0,1], with larger values indicating smaller differences between two images. As such, it can be used as a feedback signal to reflect the deformation degree of the contact surface since the value varies with the contact depth between the object and the tactile skin. Here we measure the degree of light contact by calculating the SSIM value of the real-time tactile image and the undistorted image when the sensor is not in contact with the object. 

To make the contact between the sensor skin and fragile objects light enough, as well as to obtain an effective tactile image, the ideal range of contact depth is set between 2 mm and 4 mm. Due to the uncertainty of object shape, size, and pose in the real-world test, a suitable SSIM threshold should then be set to ensure the contact depth falls within this ideal range. Therefore, we placed the sensor in contact with all objects from the 3D-printed training set with a contact depth of 3\,mm, making 50 contacts per object and randomly changing the sensor pose before each contact. The calculated SSIM value for each contact was recorded and the average value rounded to 1dp was set as the threshold (which was 0.6).

\begin{figure}[t!]
	\centering
	\begin{tabular}[b]{@{}c@{}}
        \includegraphics[width=0.8\columnwidth,trim={0 0 0 0},clip]{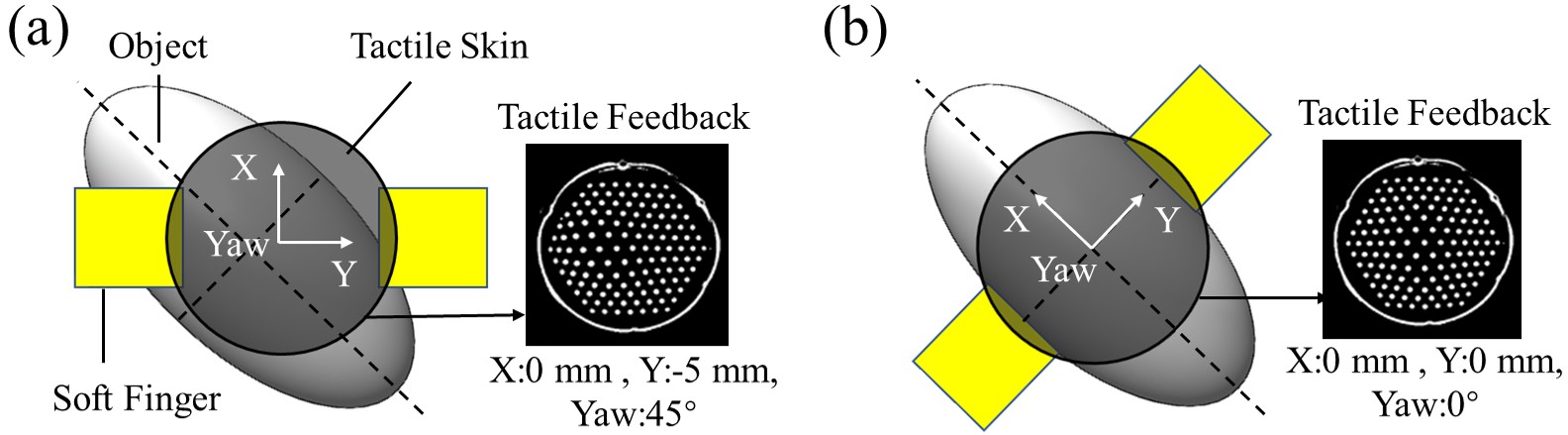} 
    \end{tabular}
	\caption{\textcolor{black}{Pose adjustment of objects with a curved surface. (a) Initial contact (overhead view). (b) Grasp pose adjustment to be central and aligned with the minor axis of the object.}}
	\label{fig:5}
 \vspace{-1em}
\end{figure} 

\begin{figure}[t!]
	\centering
	\begin{tabular}[b]{@{}c@{}}
        \includegraphics[width=1\columnwidth,trim={0 0 0 0},clip]{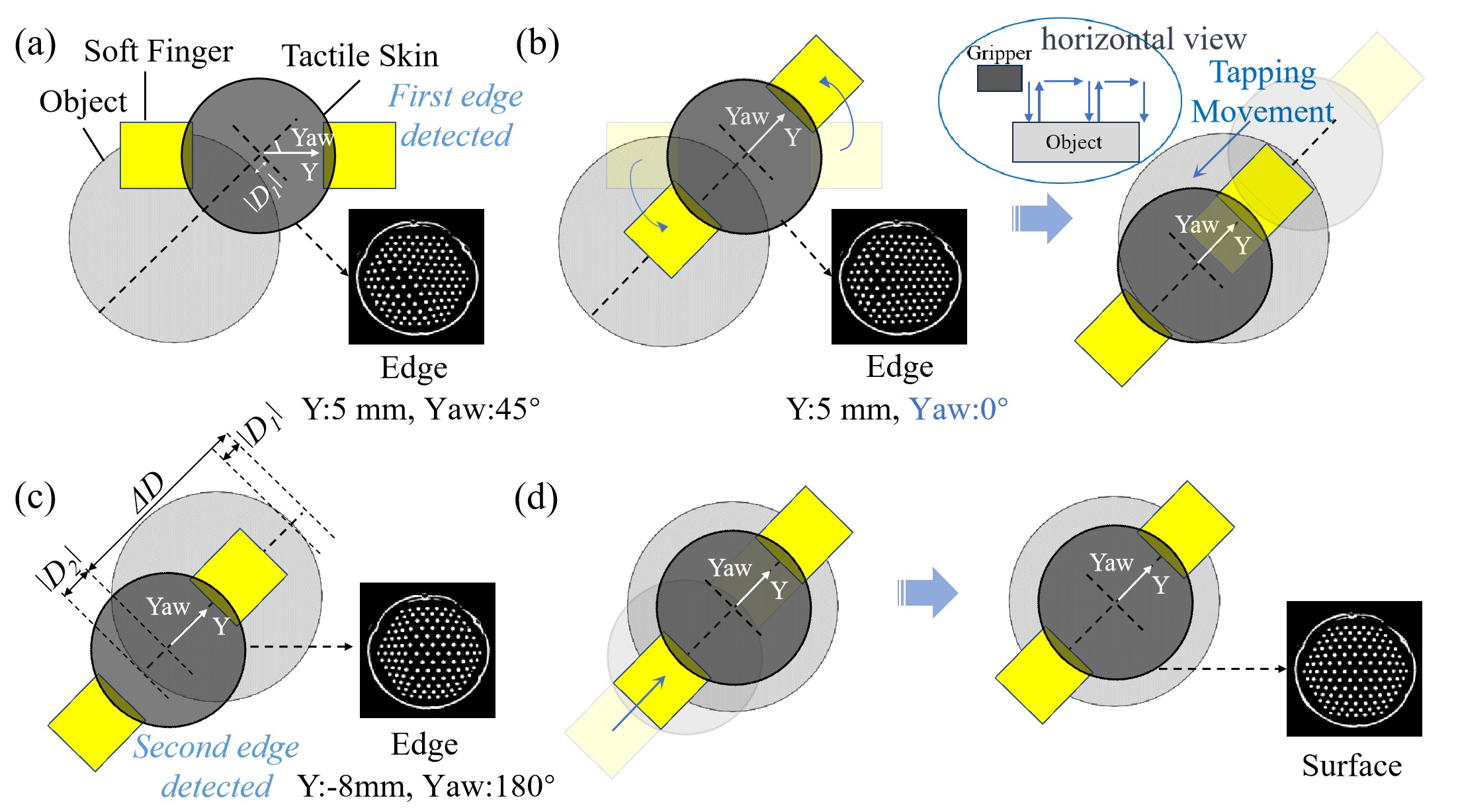} 
    \end{tabular}
	\caption{\textcolor{black}{Pose adjustment of objects with edges. (a) Initial contact (overhead view, $D_1$ is the estimated edge distance of the first edge). (b) Pose adjustment to be normal to the edge and exploration to the opposite edge. (c) Detect the opposite edge ($D_2$ is the estimated edge distance of the second edge, $\Delta D$ is the movement distance of the gripper).  (d) Grasp pose adjustment to be central to the object.}}
	\label{fig:6}
	\vspace{-1em}
\end{figure} 

\subsection{Grasp Adjustment}

\begin{figure*}[t!]
	\centering
	\begin{tabular}[b]{@{}c@{}}	\includegraphics[width=2\columnwidth]{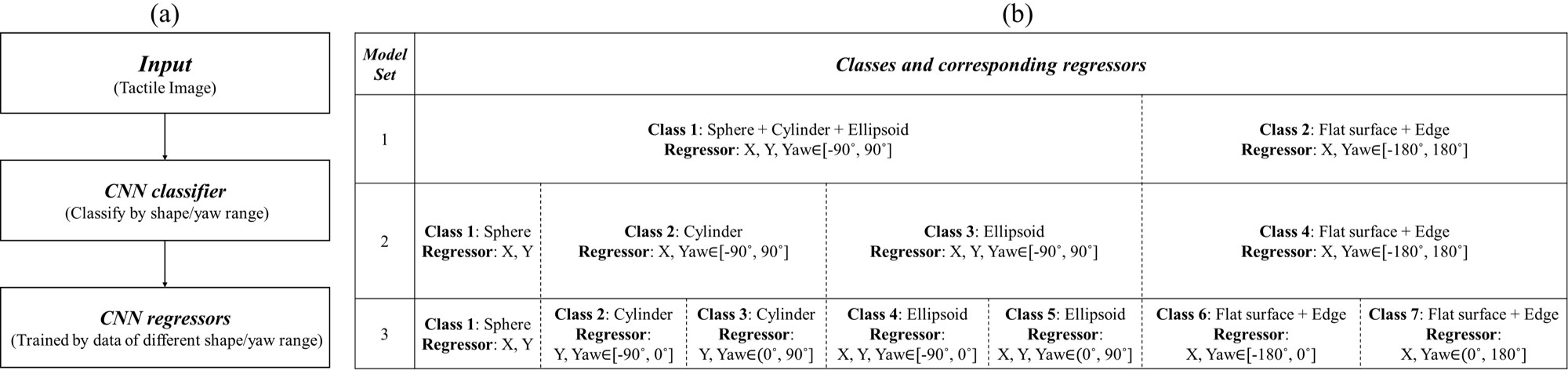} \\
	\end{tabular}
	\caption{(a) The workflow of the contact pose estimation model. (b) The proposed CNN classification and regression model sets. }
	\label{fig:7}
	\vspace{-1em}
\end{figure*}

\subsubsection{Adjustment Strategy}
The light contact detection function ends when the calculated SSIM value reaches the set threshold: the soft gripper will then adjust the grasp pose according to the tactile feedback. A trained CNN classifier is used to predict the geometric features of the contacted object based on the tactile image. Two separate pose adjustment strategies are proposed based on whether the geometric features of the object are classified as a curved surface or edge.

The adjustment strategy for the curved surface is shown in Fig. \ref{fig:5}, where a trained CNN regressor predicts the object pose, including the relative distances in the $X$ and $Y$ directions and the relative angle in the Yaw direction between the contact point and the tactile palm centre. The perceptual working field is a square area of 24$\times$24\,mm$^2$ with $X$, $Y$ ranging over $[-12,12]$\,mm around the tactile skin centre, with the perceived angle ranging within $[-90$°$,90$°$]$. The relative distances and angles are set as the postural errors, which means the optimal grasping pose is to align the contact point (which coincides with the centre of the object surface) to the centre of the tactile palm and to make the grasping direction of the parallel fingers perpendicular to the long axis of the object. The gripper executes this pose adjustment process repeatedly until the predicted distance error is less than 1\,mm and the angular error is less than 1 degree.

The adjustment strategy for an edged object is shown in Fig. \ref{fig:6}, with the relative poses predicted by a trained CNN regressor that includes the relative distance between the edge and the palm centre in the radial direction and the relative angle in the Yaw direction. The perceptual range of the relative distance $X$ is $[-12, 12]$\,mm and of the relative angle is $[-180$°$,180$°$]$. 

The CNN regressor for edge pose has a second use in also acting as a classifier of whether the palm is in contact with an edge or a flat upper surface of the object, by determining whether the predicted relative distance is outside the perceptual range of $[-12,12]$\,mm ({\em i.e.} for a flat surface, the edge is predicted to be outside the trained distance range). 

Overall, the grasp adjustment strategy on an edged object has the following steps (Fig. \ref{fig:6}):\\ 
\noindent(a) {\em Initial localization and object exploration to find the first edge}: The gripper will move downwards and contact the object. When a contact is detected, the tactile image is classified into an edge ($|X|<12$\,mm) or upper flat surface ($|X|>12$\,mm). If a flat upper surface is predicted, the gripper moves with a discrete tapping movement, classifying each tactile image until an edge is predicted. The discrete taping movement represents a repetitive motion by first moving upward 8\,mm to disengage from the object, then horizontally 24\,mm and finally downward 8\,mm to make contact with the object again (see Fig. \ref{fig:6} (b)). The sensing width of the tactile palm is $[-12,12]$\,mm; hence, the distance of each horizontal movement is set to 24\,mm. \\
\noindent(b) {\em Pose adjustment to be normal to the edge and exploration to the opposite edge:} If an edge is predicted, the relative edge pose is estimated from the CNN regressor, composed of the edge distance $D_1$ (possibly negative) and edge angle \textit{Yaw}. The gripper then adjusts its direction to be perpendicular to the edge and moves in discrete taps across the direction it is oriented (in 24\,mm steps), classifying each tactile image until another edge is again predicted. \\
\noindent(c) {\em Object width determination:} The CNN regressor estimates the distance $D_2$. This gives an object width $W = D_1$ + 24$\times$steps + $D_2$, which is the distance between the two edges.\\
\noindent(d) {\em Grasp}: The gripper moves a distance $-(D_2 + W/2)$ to the centre of the two edges and aligns its parallel fingers perpendicular to the two edges to perform the grasp.

\subsubsection{CNN Model Sets}

The workflow of the pose estimation model comprises three steps (Fig. \ref{fig:7} (a)): (1) gather the input tactile image; (2) then a CNN classifier predicts the object shape (three curved upper surfaces: sphere, ellipsoid and lateral surface of cylinder, one flat upper surface or edge for an edged object); (3) conditioned on the object shape, a CNN regressor predicts the object feature pose, including the position $(X,Y)$ and angle {\em Yaw} from the centre of the palm. We consider three different models M1-M3 built from these steps with different versions of CNN classifiers and regressors.     

The simplest version is model M1 (Fig. \ref{fig:7} (b), top row), where the shape is classified into two classes: either a curved upper surface or a flat upper surface/edge. For a curved upper surface, the CNN regressor is trained to predict {\em Yaw} between $[-90$°$,90$°$]$; for a flat upper surface/edge, the CNN regressor is trained to predict {\em Yaw} between $[-180$°$,180$°$]$. These ranges are based on considering the object symmetries: the curved upper surfaces for training are all reflection symmetric around an axis, so only a 180° {\em Yaw} range need to be considered (the other 180° corresponding to the negative distance); meanwhile, edges are distinguishable across a full 360° {\em Yaw} range.

The next model M2 (Fig. \ref{fig:7} (b), middle row) separates the objects across four classes: the sphere, ellipsoid and lateral surface of a cylinder, and a flat upper surface/edge. The reason for this is that the sphere is rotationally symmetric, so the {\em Yaw} angle is not well defined in the regressor, unlike the ellipsoid and lateral surface of the cylinder. Similarly, the relative distance $X$ is also not well defined for lateral cylinder, so these should be treated separately. Accordingly, the CNN regressor is trained to either not predict {\em Yaw} (Sphere), predict {\em Yaw} between $[-90$°$,90$°$]$ (ellipsoid and lateral surface of a cylinder), or predict between $[-180$°$,180$°$]$ (flat surface/edge). 

The last model M3 (Fig. \ref{fig:7} (b), bottom row) further separates the object orientations into positive and negative {\em Yaw} angles, treating the positive/negative classification as part of the object class (so there are 7 classes, as the sphere has no {\em Yaw} prediction). The reason is that we found an issue with {\em tactile aliasing}~\cite{lloyd2021probabilistic}, where the training shapes at the positive and negative extremes of the Yaw range can have near identical tactile images, which can lead to poor model performance when trained on aliased tactile images, as shown in the left image of Fig. \ref{fig:10}. To alleviate this issue, we perform the regression over just the positive or negative yaw range (scaled to also be positive), and then multiply this by the sign of the {\em Yaw} angle from the object classification.



\begin{figure}[b!]
	\centering
	\begin{tabular}[b]{@{}c@{}}
        \includegraphics[width=\columnwidth,trim={0 0 0 0},clip]{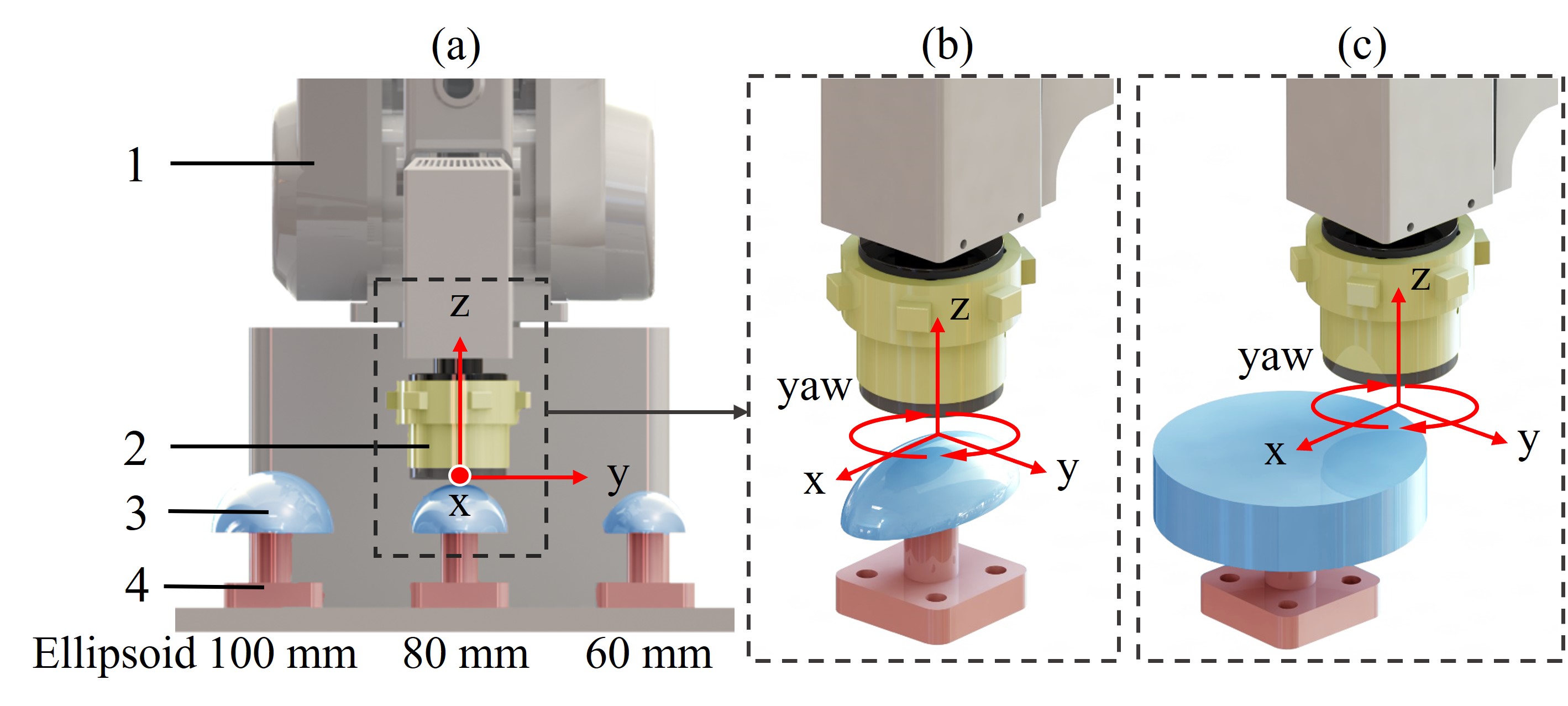} 
    \end{tabular}
	\caption{\textcolor{black}{ (a) Data collection setup. (1-Robot arm. 2-Tactile sensor. 3-3D-printed objects with different size. 4-Object base.) (b) Data collection on objects with curved surface. (c) Data collection on objects with edges. }}
	\label{fig:8}
	\vspace{-1em}
\end{figure}
\begin{table}[b!]
\centering
\caption{\label{tab:Table_with_numbers1} The range of different pose for data collection.}
 \begin{tabular}{|c | c |c |c |c |} 
 \hline
  \textbf{\textit{Shape}} &
  \textbf{\textit{\makecell{X (mm)}}} &
  \textbf{\textit{\makecell{Y (mm)}}} &
  \textbf{\textit{\makecell{Z (mm)}}} &
  \textbf{\textit{\makecell{Yaw (°)}}} \\ [0.5ex] 
 \hline
Hemisphere & [-12, 12] & [-12, 12] & [-1,1] & -- \\
Lateral cylinder & -- & [-12, 12] & [-1,1] & [-90, 90]\\
Ellipsoid & [-12, 12] & [-12, 12] & [-1,1] & [-90, 90]\\
Edged flat surface & [-12, 12] & -- & [-1,1] & [-180, 180]\\
\hline
\end{tabular}
\label{table:example}
\vspace{0em}
\end{table}

\subsubsection{Data Collection and Model Training}
We used an automated data collection method adapted from \cite{lepora2022digitac} using the test platform developed for this work. The 3D-printed objects used for data collection are connected to the object base through a square slot, and the object base is bolted to the mounting holes on the acrylic plate (Fig. \ref{fig:8}), so that the precise pose of the object relative to the robot end-effector can be obtained. 

During this data collection process, we took advantage of the modular design to remove the finger modules, so that they do not impede contact between the tactile palm and test object (we could have applied a negative pressure, but this was impractical given the large number of samples taken). 

During the acquisition process, the initial pose of the object and palm are aligned (as shown in Fig. \ref{fig:8} (b) and (c)), and then the tactile palm is controlled by the robot arm to contact the object with a random pose within the perceptual workspace (ranges in Table 1) with reference contact depth in the $Z$ direction of 3\,mm that is varied by $[-1,1]$\,mm. The relative $(X, Y)$ position, {\em Yaw} angle and tactile image are automatically recorded for each contact. 4000 tactile images are collected for each of the 12 regular training objects (white objects in Fig. \ref{fig:3}), retaining whether it has a hemi-spherical, lateral cylindrical, ellipsoidal or edged shape. In addition, 750 tactile images and object/pose labels are collected on 8 further test objects (yellow objects in Fig. \ref{fig:3}). 

For training the models M1-M3 detailed in Section~IV-C, the training data is partitioned into appropriate classes for the distinct regression models as shown in Fig. \ref{fig:7} (b). All data for regression is aggregated within the same class; in particular, for Model~M3 the negative {\em Yaw} ranges are scaled to be positive, with the sign of the {\em Yaw} angle predicted by the classifier component of the model. Model architectures, hyperparameters and training/validation methods are taken from Ref.~\cite{lepora2022digitac} and here we report the differences from previous methods: (i) Previously, the data collection was conducted on flat surfaces and straight edges, whereas our data collection involved curved surfaces of various shapes and curved edges with varying curvatures. (ii) Here we adjusted the curvature of the sensor surface, transitioning it from a hemisphere to a surface with a radius of 42\,mm. This modification allowed for effective deformation upon contact with both curved and flat surfaces, ensuring the capture of valuable tactile information. Our CNN network comprises 5 convolutional layers and 1 dense layer, utilizing ELU and ReLU activation functions for the surface and edge models, respectively. The training and validation process employed an 80/20 split, spanning 100 epochs with early stopping implemented after 10 epochs. The model was trained on an NVIDIA MX550 GPU, with an average training time of 40 minutes for model set 1 to 3.

\subsection{Loss-of-contact Detection}
After the pose adjustment is completed, the object will be grasped and lifted. The object will tend to detach under its own gravity or when the grasping force fluctuates after lifting, which results in a reduction of its contact depth with the tactile skin. Loss-of-contact detection can be regarded as the inverse process of contact detection from Section IV.B, using the SSIM value to measure the disengagement from the tactile skin at the palm. To set the threshold, we used the soft gripper to perform light contact detection and the pose adjustment on each object in the 3D-printed training set, with the contact depth set to 3\,mm. The calculated SSIM value was recorded when the pose adjustment was completed but before the object was lifted. The process was repeated 100 times for each object and the average SSIM value was set as the threshold (which was 0.66). The air pump activates when loss of contact is detected, increasing finger air pressure. As the fingers flex more, they apply greater force to the object, pressing it into the tactile palm. The air pump stops pressurizing when object detachment is no longer detected. To safeguard fragile objects like eggs and glass, we avoid applying a strong initial grasping force. Instead, we rely on the loss-of-contact detection function. This approach begins with a gentle initial grasp and gradually increases the force through closed-loop control, ensuring maximum protection against object damage.

\section{Experiment Results}

\subsection{The Maximum Contact Depth Analysis}
\begin{figure}[b!]
	\centering
	\begin{tabular}[b]{@{}c@{}}
 \includegraphics[width=0.8\columnwidth,trim={0 0 0 0},clip]{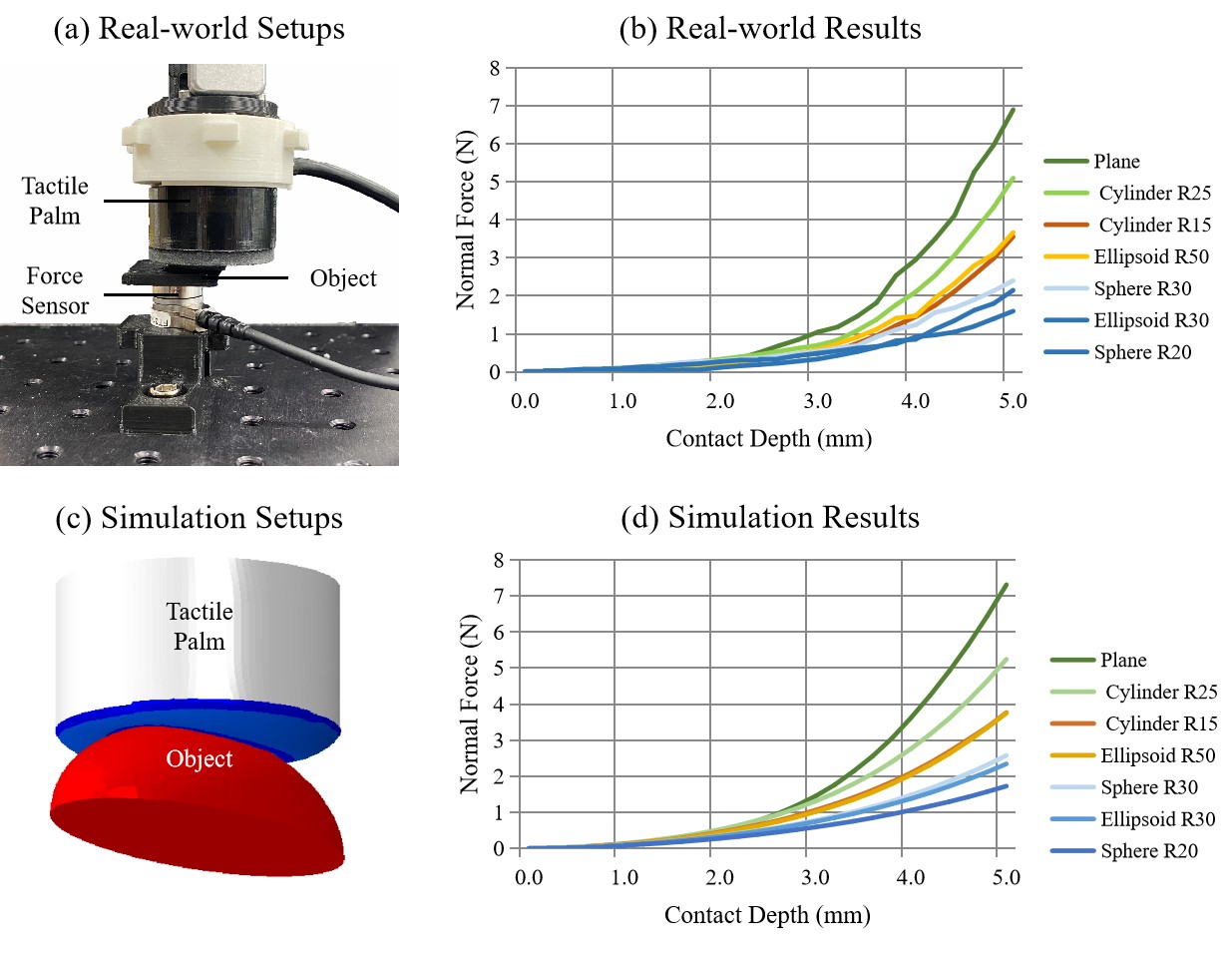} 
    \end{tabular}
	\caption{\textcolor{black}{The normal contact force test in both real-world and simulation environments. The contact is slight enough to protect the delicate objects.} }
	\label{fig:9}
	\vspace{-1em}
\end{figure}

We investigated the relationship between contact depth and normal contact force in both real-world and simulated environments to determine the maximum contact depth achievable between the palm and the fragile objects without damaging them. In the simulated environment, we modelled the normal contact between objects of various shapes and the tactile palm, recording the normal contact force curves as the compression depth increased. In the real world, we used a Nano 17 force/torque sensor placed on the compressed object to measure the normal force during contact, as illustrated in Fig. \ref{fig:9} (a,c). The results demonstrate that the maximum normal force is less than 8N when the contact depth reaches 5 mm, whereas common fragile objects, such as eggs, can typically endure forces up to 30 N, which is much higher than the contact forces observed in the tests. Therefore, when the contact depth reaches 5 mm, the generated force is sufficiently low to effectively protect the object from damage while providing valid tactile feedback. Moreover, the membrane of our tactile palm is 5 mm thick, limiting the maximum contact depth to 5 mm. We've included a 1 mm compression margin for sensor protection, setting the maximum compression depth at 4 mm.

\subsection{Offline Model Analysis}
The generalization capability of the proposed pose estimation CNN model set is tested on all the objects in the test 3D-printed object set (objects coloured yellow in Fig. \ref{fig:3}). Each object is tested with 750 random contacts, comparing the predicted pose and ground-truth pose, with the gripper initialized in a random pose before contacting. The Mean Average Error (MAE) is used as an evaluation metric to select the best CNN model set for the subsequent online experiments. The test results are presented such that data with minimal errors are highlighted in bold (Table~2). 

For relative distance prediction, Model Set 3 has the smallest prediction error on all tested objects, although all model sets achieve accurate (sub-millimeter) prediction accuracy on objects other than the ellipsoid. 

For relative angle prediction, the prediction errors of these model sets varied widely. Overall, Model set 3 had the best prediction error, achieving a prediction accuracy that was less than 2 degrees on all objects except for the ellipsoid (2.35 degrees). The other model sets performed far more poorly, typically being over 10 degrees mean error. 

\begin{figure}[t!]
	\centering
	\begin{tabular}[b]{@{}c@{}}
        \includegraphics[width=\columnwidth,trim={0 265 0 22},clip]{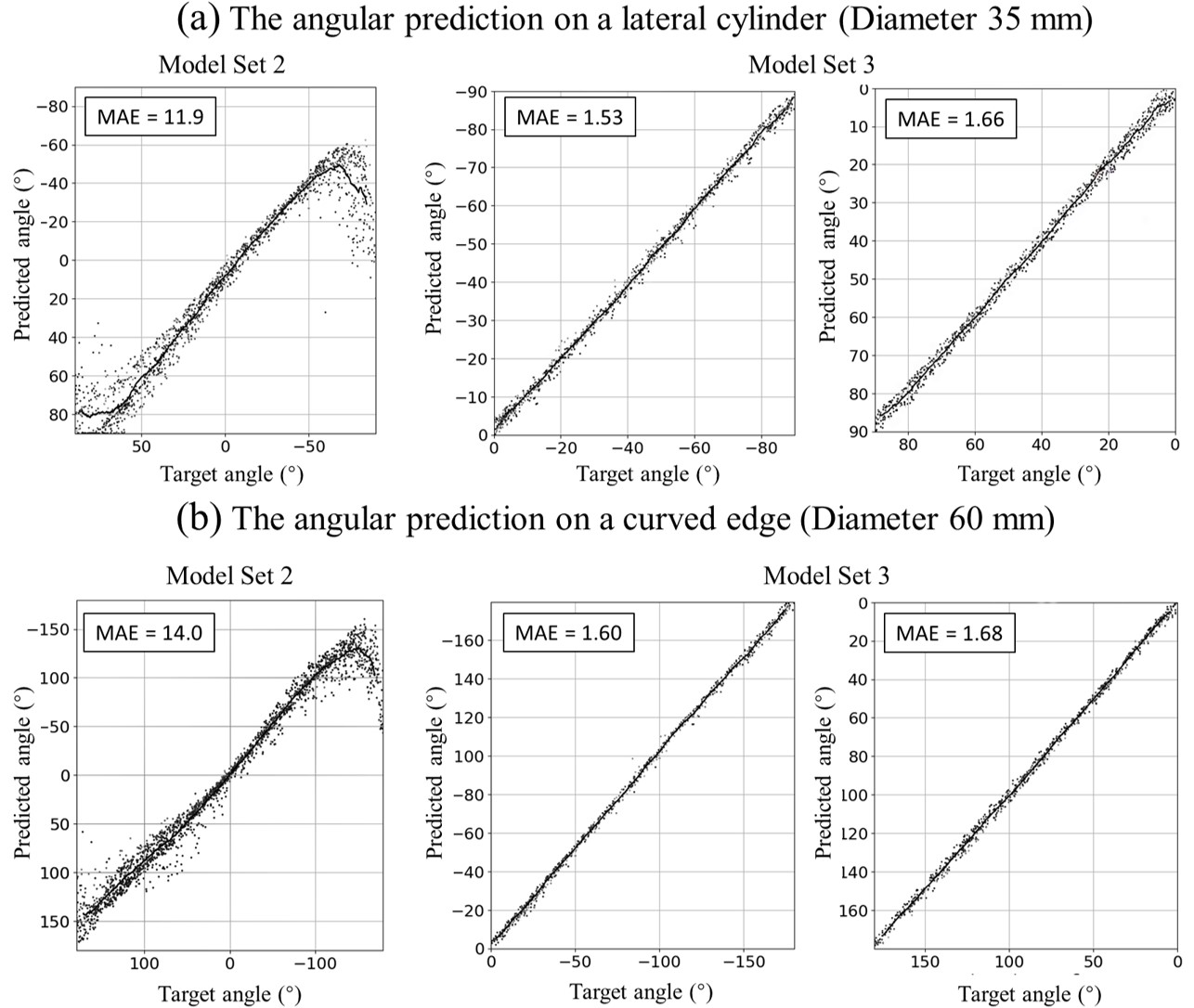} 
    \end{tabular}
	\caption{\textcolor{black}{The angular prediction performance of CNN model sets 2 and 3 (see Fig. 7(b)) on a cylinder oriented laterally (diameter 35 mm) and their MAEs.}}
	\label{fig:10}
	\vspace{-1em}
\end{figure}

To examine the cause of this difference in angle performance over the models, we display the test results on a lateral cylinder (diameter 35 mm) and a curved edge (diameter 60 mm) in Fig. \ref{fig:10} for model sets 2 and 3. For model set 2, there is a clear deviation of the predictions from the ground truth at the extremes of the angular range ($\pm90^\circ$ for the lateral cylinder and $\pm180^\circ$ for the curved edge) that is not present in model set 3. We interpret the issue as being a {\em tactile aliasing} problem that was studied recently in ref.~\cite{lloyd2021probabilistic}: because of the symmetry of the object, tactile data from the cylinder is similar at $\pm90^\circ$. As a consequence, angle prediction near those extremes should be bi-modal; however, the training of standard (non-probabilistic) neural networks assumes a uni-modal distribution. As a consequence, the predictions smoothly tend to the mean of $\pm90^\circ$, i.e. near zero, which produces regions of poor predictions for model set 2. In ref.~\cite{lloyd2021probabilistic}, the solution was to use a neural network that represents multi-modal probability distributions; here we use a simpler method of separating the angle ranges so no single model within model set 3 suffers from tactile aliasing. 

\begin{table}[t!]
\centering
\caption{\label{tab:Table_with_numbers2}Test results of different CNN model sets on 3D-printed testing sets.}
 \begin{tabular}{|c|c|c|c|c|c|} 
 \hline
  \textbf{\textit{Object}} &
  \textbf{\textit{\makecell{Diameter\\(mm)}}} &
  \textbf{\textit{\makecell{Model\\Set}}}&
  \textbf{\textit{\makecell{MAE\\$X$(mm)}}} &
  \textbf{\textit{\makecell{MAE\\$Y$(mm)}}} &
  \textbf{\textit{\makecell{MAE\\{\em Yaw}(°)}}} 
 \\ [0.5ex] 
 \hline
 \hline
\multirow{6}*{Hemisphere} & \multirow{3}*{45} & M1 & 0.64 & 0.6 & \multirow{6}*{--} \\
 &  & M2 & 0.52 & 0.25 & \\
 &  & M3 & \textbf{0.51} & \textbf{0.23} &  \\
\cline{2-5}
 & \multirow{4}*{55} & M1 & 0.8 & 0.98 &  \\
 &  & M2 & 0.56 & 0.9 &  \\
 &  & M3 & \textbf{0.54} & \textbf{0.88} &  \\
\hline
\multirow{6}*{Lateral cylinder} & \multirow{3}*{35} & M1 & \multirow{6}*{--} & 0.87 & 9.84 \\
 &  & M2 &  & 0.80 & 11.9 \\
 &  & M3 &  & \textbf{0.23} & \textbf{1.66} \\
\cline{2-3}  \cline{5-6}
 & \multirow{3}*{45} & M1 &  & 0.73 & 9.10 \\
 &  & M2 &  & 0.97 & 14.4 \\
 &  & M3 &  & \textbf{0.29} & \textbf{1.45} \\
\cline{1-6}
\multirow{6}*{Ellipsoid} & \multirow{3}*{70} & M1 & 1.00 & 0.62 & 7.36 \\
 &  & M2 & 1.79 & 0.98 & 14.6 \\
 &  & M3 & \textbf{0.64} & \textbf{0.26} & \textbf{1.73} \\
\cline{2-6}
 & \multirow{3}*{90} & M1 & 1.52 & 0.97 & 15.2 \\
 &  & M2 & 0.92 & 1.89 & 37.9 \\
 &  & M3 & \textbf{0.82} & \textbf{0.27} & \textbf{2.35} \\
\cline{1-6}
\multirow{4}*{Curved edged} & \multirow{2}*{60} & M1,M2 & 0.5 & \multirow{4}*{--} & 14.0 \\
 &  & M3 & \textbf{0.46} & & \textbf{1.68} \\
\cline{2-4} \cline{6-6}
 & \multirow{2}*{80} & M1,M2 & 0.32 &  & 16.6 \\
 &  & M3 & \textbf{0.29} & & \textbf{1.59} \\
\hline
\end{tabular}
\label{table:example1}
\vspace{-1em}
\end{table}

Based on this method, we use data collected from the same shaped objects to train the corresponding CNN regression models, combined with a CNN classifier for shape classification of the tactile images, which obtains the highest pose prediction accuracy. Therefore, model set 3 is used as the optimal model for subsequent experiments.

\subsection{Tactile Grasping Tests}
\begin{figure*}[t!]
	\centering
	\begin{tabular}[b]{@{}c@{}}	\includegraphics[width=1.3\columnwidth]{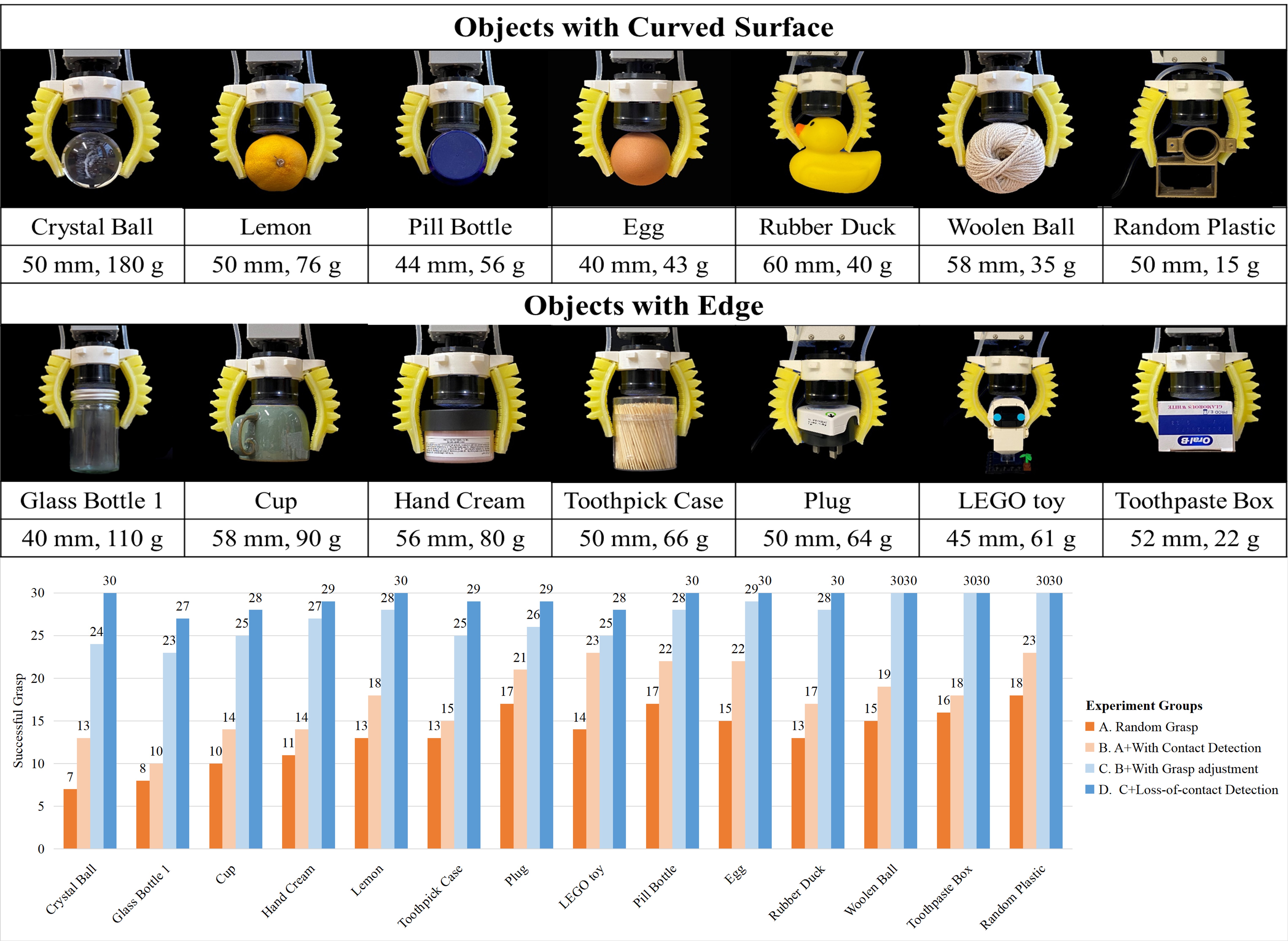} \\
	\end{tabular}
	\caption{Top: everyday objects used for grasping.  Bottom: grasping stability results comparison. }
	\label{fig:13}
	\vspace{-0.5em}
\end{figure*}

A wide range of objects was used for the soft tactile grasping strategy tests, including all objects in the 3D-printed test set and several fragile everyday objects, as shown in Fig. \ref{fig:13} (left). To verify the grasping strategies, we fix the initial pose of the object and randomize the initial pose of the soft gripper in each test. 

\subsubsection{Light Contact Detection Test and Loss-of-contact Detection Test}
For the light contact test, we placed the tested object in a random pose under the soft gripper and kept the contact surface within the perceptual workspace of the tactile palm. The soft gripper was controlled by the robotic arm to move slowly downward with a fixed pose until the calculated SSIM reached the set threshold, then the contact depth at this time was recorded. The contact depth was recorded as a negative value if the object and the tactile skin were not in contact. Each object was tested 20 times.
 
\begin{figure}[t!]
	\centering
	\begin{tabular}[b]{@{}c@{}}
        \includegraphics[width=0.9\columnwidth,trim={0 0 0 0},clip]{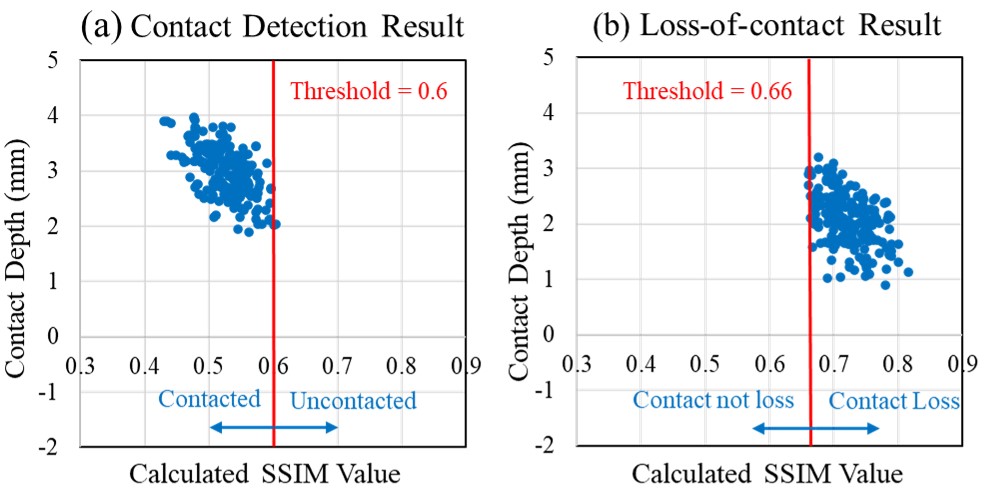} 
    \end{tabular}
	\caption{\textcolor{black}{Test Results of contact detection and loss-of-contact detection on a 3D-printed test set and household objects.}}
	\label{fig:11}
	\vspace{-1em}
\end{figure}

The results show that the contact depths recorded in the tests were all positive (Fig. \ref{fig:11} (a)), indicating that contact was detected in every test; i.e. there were no false positive detections of contact. We observed that the objects did not move during contact with the tactile palm regardless of their initial pose, suggesting that the contact was light enough to protect fragile objects. The recorded contact depths were all within the ideal contact range from 2-4\,mm, with an average depth of 2.9 mm, which indicates that the threshold for light contact was set to a reasonable level.

For the loss-of-contact test, after performing light contact detection and pose adjustment, we artificially pulled the object down gently in the Z-direction after it was lifted, then recorded the calculated SSIM and the corresponding contact depth after the loss-of-contact was detected. The test results show that all recorded contact depths decreased in comparison but were still positive with an average value of 1.8\,mm (Fig. \ref{fig:11} (b)), again indicating that all loss-of-contacts were detected reliably with no false positives.

\begin{figure}[t!]
	\centering
	\begin{tabular}[b]{@{}c@{}}
        \footnotesize{\hspace{2.5em}(a) Egg\hspace{12em} (b) Bottle}\\
        \includegraphics[width=0.9\columnwidth,trim={0 0 0 22},clip]{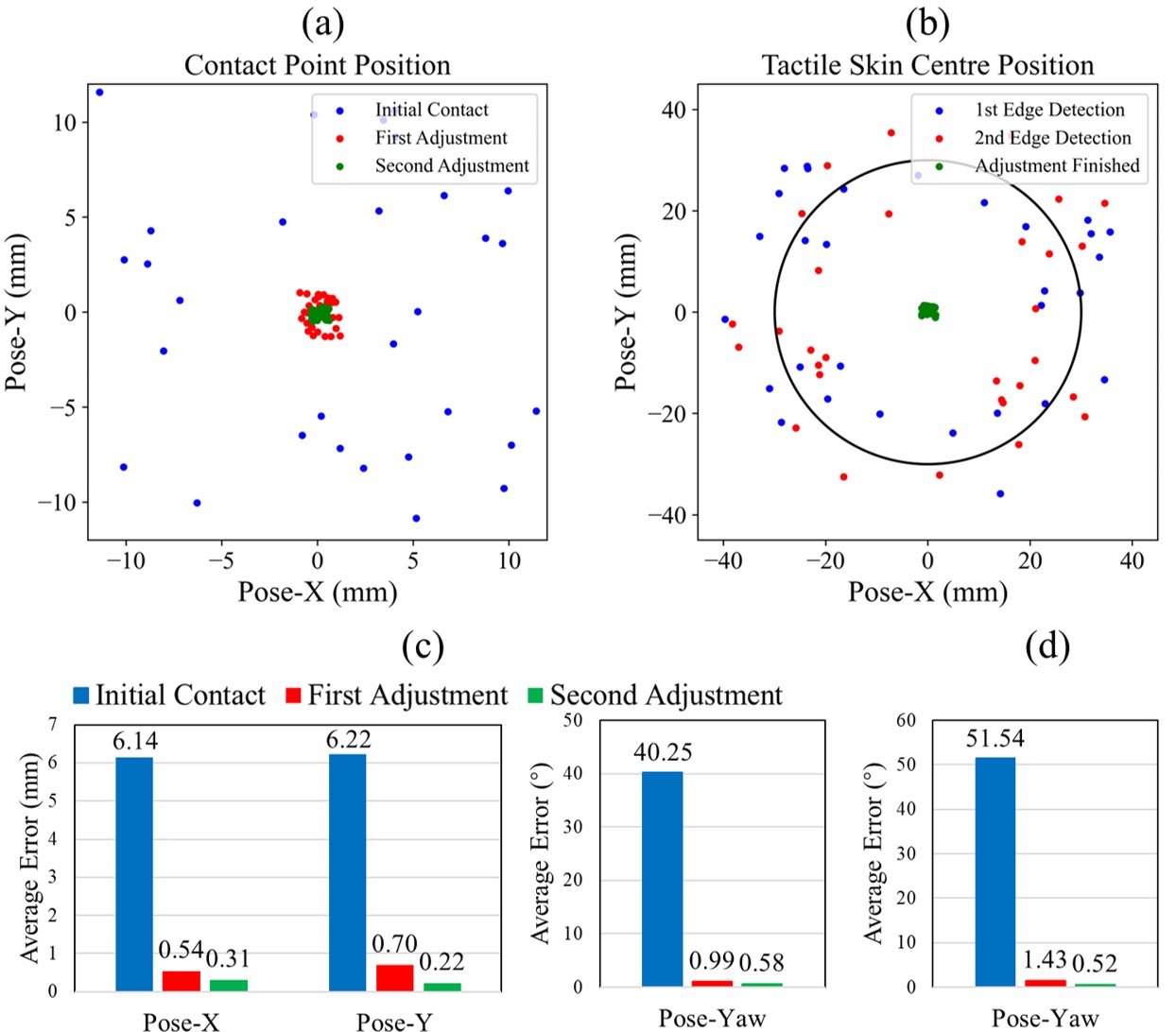} 
    \end{tabular}
	\caption{\textcolor{black}{Test results of grasp pose adjustment on an egg (left panels) and a glass bottle (right panels). (a,c) In (a), the entire square represents the 12mm x 12mm tactile sensing area of the palm, with the dot positions indicating the distribution of the contact center within this sensing area. For the egg, contact positions within the set perception area for curved objects and their average error during grasping adjustment. (b,d) In (b), the circle at central position represents the circular edge of the cylindrical glass bottle, while the dots indicate the center positions of tactile palms. For the bottle, positions of the tactile palm centre relative to the detected edge and the average angular error during the angle adjustment.}}
	\label{fig:12}
	\vspace{-1.5em}
\end{figure}

\subsubsection{Grasping Pose Adjustment Test}

For this test, we examine the pose adjustment method following the light contact detection stage. For objects with curved surfaces, we set the pose adjustment to be performed until the relative distance error is less than 0.5 mm, and the relative angle error is less than 1 degree, with the relative poses of each adjustment recorded. For objects with edges, the gripper only adjusts its pose after the first edge is detected, and the adjustment stops when the relative angular error is less than 1 degree. The position of the tactile palm centre is recorded whenever an edge is detected and when the pose adjustment is complete. Thirty tests were performed on each object, with the average error of the relative pose calculated and recorded.

For the first example, the results of the 30 grasp adjustment tests on eggs are shown in Fig. \ref{fig:12} (a) and (c). In Fig. \ref{fig:12} (a), the entire square represents the 12 mm $\times$ 12 mm tactile sensing area of the palm, with the dot positions indicating the distribution of the contact center within this sensing area. The initial contact on the egg was distributed randomly within the perceptual workspace of the tactile palm, with average relative distances of 6.1\,mm, 6.2\,mm in the X-, Y-directions, respectively, and an average relative Yaw angle of 40.2 degrees. After the first adjustment, the pose of the soft gripper was similar to that of the egg, with an average distance error less than 1\,mm and angular error less than 1 degree, indicating that the gripper had reached the target pose after the first adjustment in most tests. The error was further decreased after the second adjustment, with all errors reaching the target threshold, thus the contact centres will further converge towards the center of the sensing area.

As a second example, the results of the cylindrical glass bottle are shown in Fig. \ref{fig:12} (b) and (d). In Fig. \ref{fig:12} (b), the circle at central position represents the circular edge of the cylindrical glass bottle, while the dots indicate the center positions of tactile palms. Initially, the sensor positions are randomly distributed around the circular edges of the bottle. Following the adjustment strategy shown in Fig. \ref{fig:6}, when the second edge is detected, the sensor's position is located on the opposite side of the bottle's circular edge compared to its position when the first edge is detected, resulting in similar distributions. The perception range of the relative edge angle is (-180°, 180°), but because of the gripper's 180° rotational symmetry, we can choose an adjustment angle less than 90° (technically $\arctan(\tan\rm Yaw))$) for more efficient adjustment. The results show that the angular error is less than the 1 degree threshold after adjustment, and the tactile palm successfully detected both edges and moved to the geometric center of the object in all tests.

\begin{figure*}[t!]
	\centering
	\begin{tabular}[b]{@{}c@{}}	\includegraphics[width=2\columnwidth]{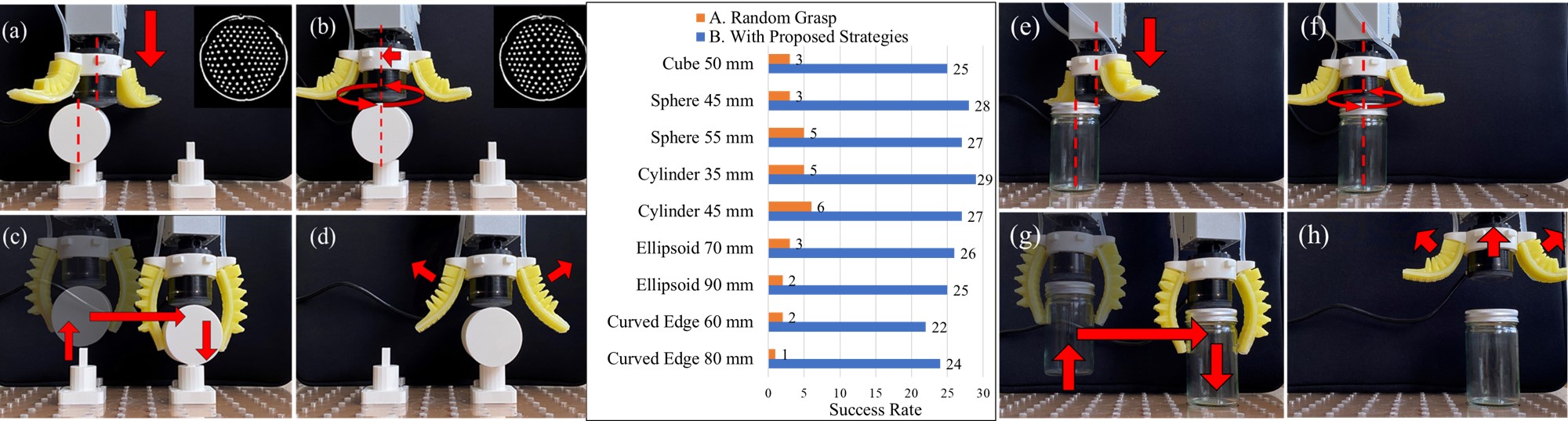} \\
	\end{tabular}
	\caption{Placement precision test: (a) Contacting and Sensing. (b) pose adjustment. (c) Grasping and moving. (d) Releasing.}
	\label{fig:14}
	\vspace{-1em}
\end{figure*}

\subsection{Grasp Stability Test}

To verify the enhancement of grasping stability by the proposed tactile strategies of contact detection, grasp pose adjustment and loss-of-contact detection, we set up four experiment groups. The baseline experimental group considered random grasping, using the palm part as a passive support without any tactile feedback. Then the other three groups add sequentially the contact detection, pose adjustment and loss-of-contact detection tactile capabilities. In each test, objects were placed within the perceptual workspace below the gripper in random poses with the initial pose of the gripper fixed. Each object was grasped 30 times with the grasp considered successful if the object did not fall within 5 seconds after being grasped. 
In the baseline experiment group, the gripper's height during grasping was randomly set within a range of $(-5, 5)$\,mm above the object because the object's position was unknown in this group, unlike the other three groups. With the incorporation of contact detection, the gripper can now sense the vertical position of the object and initiate a grasping action upon contact detection. In the pose adjustment group, the gripper goes a step further by fine-tuning the grasping position in the horizontal direction (x, y), and yaw angle, all while maintaining an unchanged vertical grasping position.

The grasping test shows successively improving performance as more tactile capabilities are included (Fig. \ref{fig:13}, bottom bar chart). For the random grasping group, the overall grasp success rate was 45\% and the poorest of the four experiment groups. We observed that the grasping success rate gradually increases as the object's mass decreases, which appears to be because of a higher tolerance to position errors for lightweight objects (under constant grasping force). Moreover, objects with curved surfaces were easier to grasp than those with edges, which we attribute to the contact area between the inflated flexible fingers and the grasping surface being larger than that of edges and flat surface, giving a more stable wrapping grasp. We observed several failure modes for grasping: firstly, the soft fingers can push the object away during the closing process, such as with rollable objects; secondly, when the object is lifted, the deformation of the soft fingers can make the force on the object become uneven, leading to the object falling from the gripper when the initial relative pose between the object and gripper differs greatly; and, in some cases, differences in air pressure across the soft fingers can lead to a reduction in grasping force, which causes the object to slide out of the grip.

With the light contact detection and random grasping, the overall success rate improved to 59\%, with most improvement on objects with curved surfaces rather than those with edges. Contact detection enabled support from the soft palm without changing the object pose, forming a stable closing force together with the closed fingers. We interpret the improvement on curved objects as due to the closed fingers being more likely to press curved objects towards the tactile palm than objects with edges,  making the grasp more stable. The uneven force on the object after lifting and differences in air pressure across fingers appear to be the main reasons for grasp failure.

With the inclusion of grasp pose adjustment, the overall success rate reached 90\% (and 100\% for objects lighter than 35g). The consistency of the pose between the object and the gripper ensured sufficient contact area between the soft fingers and the object surface, so that objects were no longer pushed away by the closing fingers, with little further deformation of fingers after lifting the objects. Objects with edges, such as cuboids, still have a lower success rate than curved objects, which we attribute to the fingers being able to wrap underneath the curved objects so less grasping force is needed. Such differences in required grasping force appear the main cause of failure, especially for objects with large masses.

With the loss-of-contact detection so that all tactile functions are used, the success rate reached 97\%. Now, objects are grasped tightly by an increased grasping force when the objects are about to disengage from the tactile palm under their gravity. Most of the grasp failures occur on objects with edges, as these objects cannot always be stably pressed against the tactile palm due to the lack of upward support from not being able to wrap the fingers underneath the object.

When grasping objects with curved surfaces like spheres, where no gripper angle adjustments are needed, the average time from contact to successful grasping is 6 seconds. This time may extend by 0 to 3 seconds if angle adjustments are required, as is the case with ellipsoids or cylinders. For objects with distinct edge features, the average grasping time is approximately 20 seconds, varying with the object's size. It is worth noting that the time spent can be reduced if we increase the speed of the robotic arm's movements.

\subsection{Placement Precision Test}

Finally, we consider a test to validate another benefit of a tactile palm with a soft gripper: to precisely place an object after being held.

To test the performance of the full tactile grasping strategy on precise placement, we set up two experimental groups with and without tactile feedback on both 3D-printed test sets and everyday objects. The settings for the random grasping group and the group with all the proposed strategies are the same as those in the grasping stability test. For the 3D-printed objects, we set up an assembly task (Fig. \ref{fig:14}, left) where the soft gripper needs to manipulate the object to transfer it from one object base to another. There is a square protrusion on the base (dimension 7\,mm) for placing the object of size 1\,mm smaller than that of the slot on the objects. The initial pose of the soft gripper is random for each grasp, and the relative positions of the two object bases are fixed and known. For the non-tactile experiments, only trials in which the object was successfully grasped were recorded.

The results show the success rate of precise placement is majorly improved with the proposed grasping strategy, from 11\% to 86\% (Fig. \ref{fig:14}, right). We interpret this improvement as due to the precise pose estimation and closed-loop pose adjustment. Most failure cases in the tactile experiments were on edged objects, since the object pose can change slightly during the lifting and moving process due to the lack of upward support and differences in grasping force. 

To verify the ability to grasp and place fragile items, we conducted a similar test on some additional fragile items in everyday life. The results are shown in Fig. \ref{fig:14} (e)-(h), from which we make several observations: (a) after the height and position of the glass bottle are detected through light contact detection, the bottle remains still during the process, indicating an appropriately light stable grasp; (b) the stable and soft grasp ensures the safety of fragile items from falling during the manipulation, with the bottle accurately placed at the target location with the same pose. Overall, the task is accomplished when the soft gripper releases the object at the appropriate location.

\section{Discussion and Future Work}
In this study, we demonstrated the benefits of a soft biomimetic optical tactile palm for a pneumatic soft gripper. We verified the improvements from several distinct tactile grasping strategies, encompassing contact detection, grasp pose adjustment and loss-of-contact detection/correction, on both grasping stability and placement precision by testing on both 3D-printed test objects and multiple everyday objects. 

The study validates the generalization of the pose estimation model for placing the grasp and contact-based grasping strategy on various objects, showcasing the feasibility of integrating camera-based optical tactile sensors on the palm. This approach suggests a new way to incorporate tactile perception in soft grippers. It also offers a versatile design that can be adapted to other soft gripper types, including cable-driven and rigid grippers. Parameters such as shape, size, acuity and morphology of the tactile skin can be customized according to the task requirements. The proposed soft grasping strategy achieves high-stability grasping and high-precision placement through simple adjustment strategies while achieving safe interactions with objects. Placing soft tactile sensors on the palm, rather than the fingers, reduces sensor quantity and offers an adaptive support surface for richer tactile data. This can provide stable object handling during lifting and movement to ensure precise positioning without altering the object's posture.

Current limitations of the soft gripper design include that it performs better when gripping harder objects compared to softer objects, as softer objects can cause more deformation during contact between the palm and the object. This deformation hinders the tactile palm's ability to precisely identify local geometric features, leading to increased pose estimation errors. The proposed soft grasping strategies are limited in their applicability to certain object shapes, currently unable to cover some special shapes such as torus-like shaped objects and complex asymmetrical objects. Additionally, the gripper's capacity to grasp objects is constrained by finger length; when the object is smaller than the finger length, it can result in interference as the finger may touch the surface on which the object is resting. 

Some directions for future work include optimizing the finger structure for precise and adaptive closed-loop control of grasping force and exploring the possibility of integrating the proposed tactile palm into other soft grippers, such as cable-driven soft grippers. Moreover, the grasp placement strategy could be integrated with a fully 6-DOF pose control within a 3D workspace. Such strategies could leverage advances in controlling robots with soft biomimetic optical tactile sensing, such as pose-based servo control~\cite{lepora2021pose} and tactile pushing manipulation~\cite{lloyd2021goal} and can be expanded to more complex scenarios. While we focussed here on parallel-finger grippers to illustrate the benefits of tactile sensing in the palm, some shapes are then challenging to grasp such as spheres and standing cylindrical objects. In future work, multi-fingered gripper structures and other changes to the gripper morphology could be explored; however, we expect that such changes would then mean that the two-fingered soft grasping control strategies from a tactile palm would need to adapt to the gripper morphology. Moreover, we will consider integrating vision and tactile sensing to enhance system performance. Vision can provide robust global information to generate an initial grasp pose and quickly move the gripper close to the object. The proposed tactile strategy can then obtain the precise pose of the object for stable grasping and movement, and then external vision can provide accurate feedback on the object's placement position. Another interesting direction is exploring the combined perception and grasping strategies that integrate tactile feedback from both the fingers and the palm. The tactile palm can compensate for the fingertips' inability to accurately estimate the initial pose of the object before grasping, while the fingertips can provide dynamic grasping feedback that the palm cannot offer during the grasping process, while the combination of both ensures the reliability of subsequent handling processes. To facilitate these developments, we will release a TacPalm repository, including the CAD model, fabrication instructions and the operation code so that the research is fully accessible for others to build upon.

\appendices


\section{Extend the tactile palm to other morphologies}

\begin{figure}[h]
	\centering
	\begin{tabular}[b]{@{}c@{}}
        \includegraphics[width=0.7\columnwidth,trim={0 0 0 0},clip]{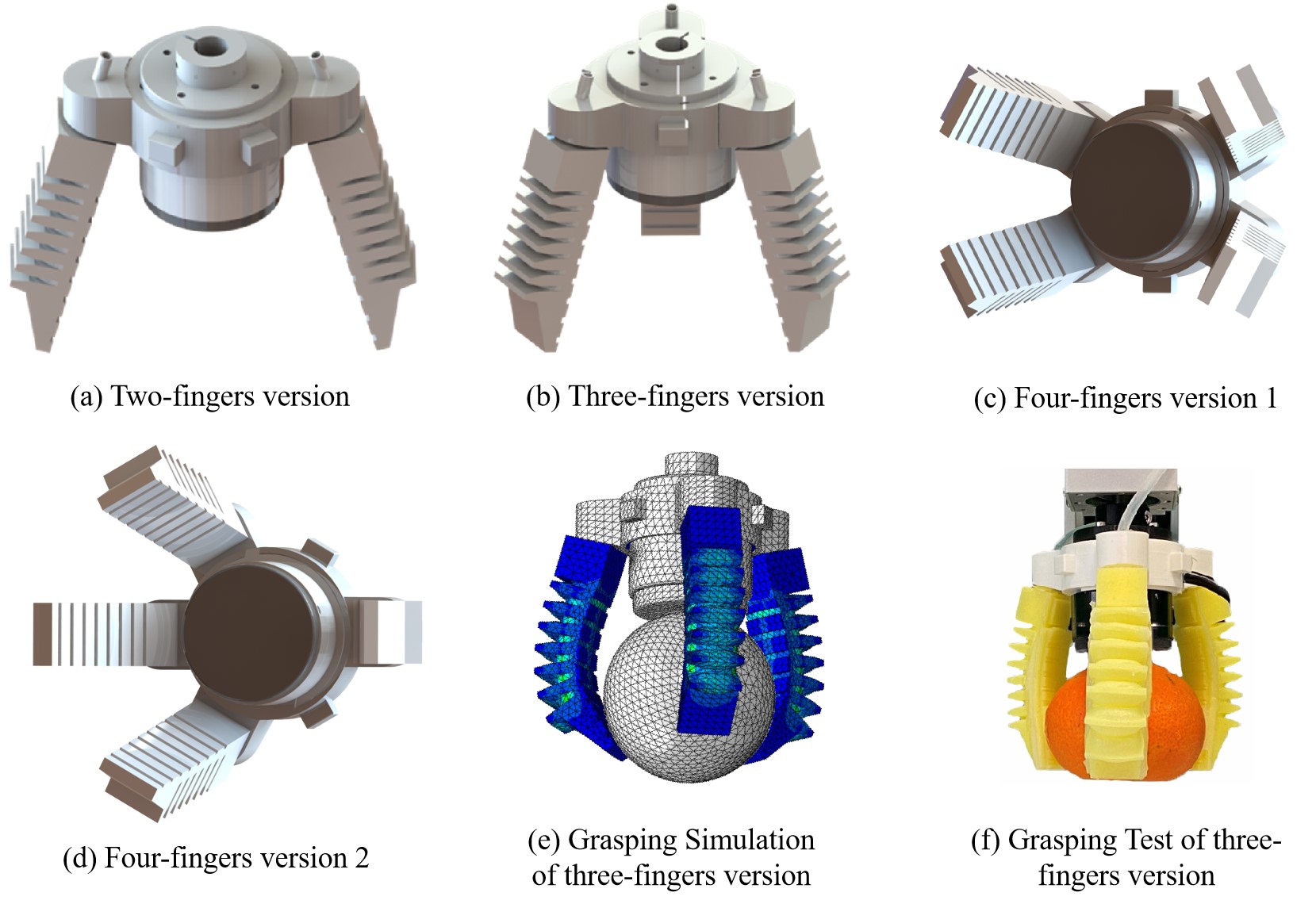} 
    \end{tabular}
	\caption{\textcolor{black}{Integration of the tactile palm into different morphologies.}}
	\label{fig:17}
	\vspace{-1em}
\end{figure}

The proposed tactile palm can be integrated into various soft morphologies due to its modular design, which allows for independent operation of the palm and fingers. Structurally, it can be connected to different numbers of fingers or different finger structures of fingers through slots on the rigid palm connection base. Its size and shape are customizable to meet different dimensional requirements. Electrically, the tactile palm operates independently of the finger drivers, using a single USB interface for power. Additionally, the tactile adjustment strategy primarily relies on feedback from the tactile palm, ensuring that changes in the finger design do not impact its effectiveness. Fig. \ref{fig:17} presents the integration of tactile sensors across different numbers of fingers, along with the corresponding results from grasping simulations and real-world grasping experiments.

\begin{figure}[h]
	\centering
	\begin{tabular}[b]{@{}c@{}}
        \includegraphics[width=0.5\columnwidth,trim={0 0 0 0},clip]{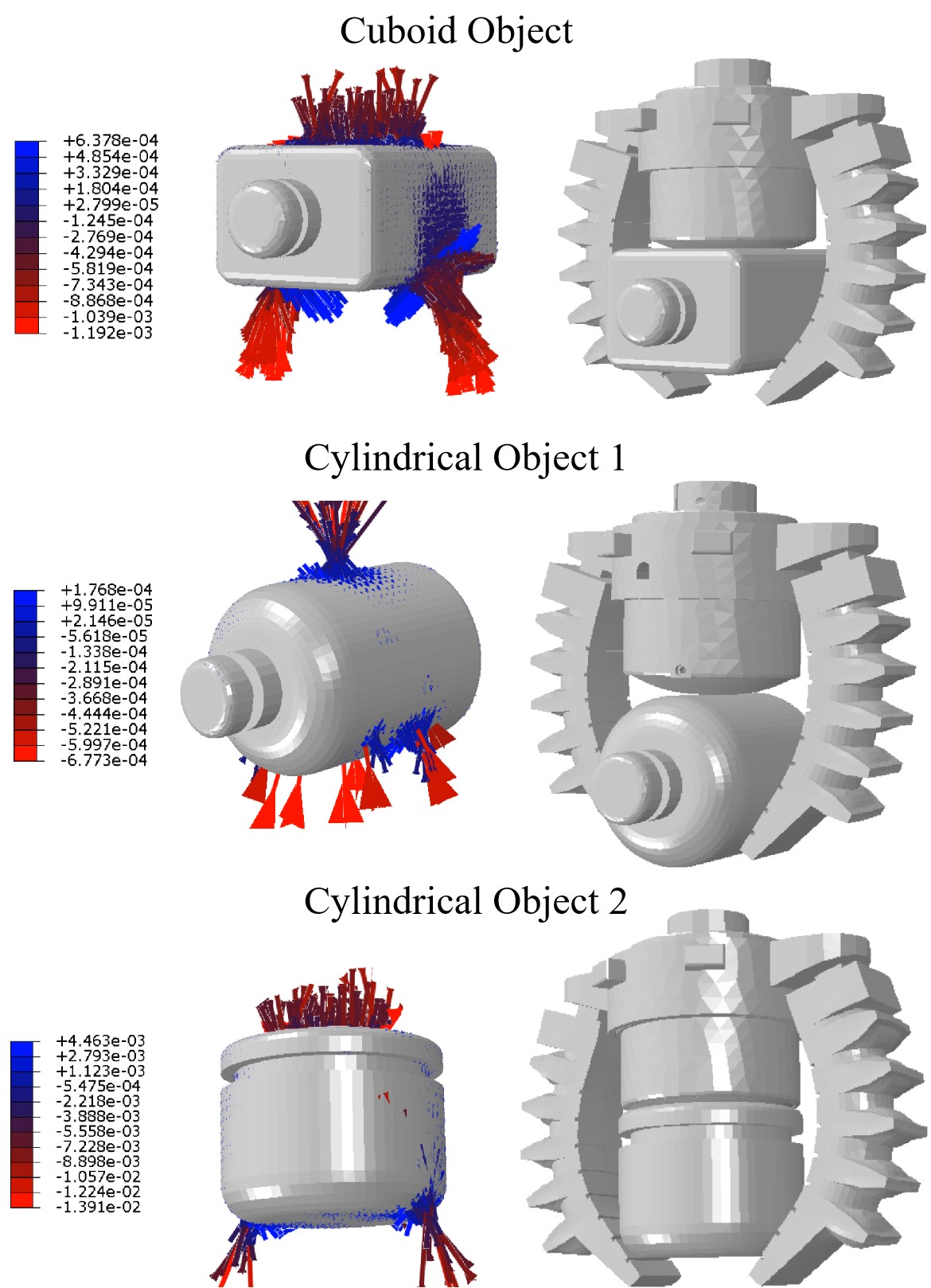} 
    \end{tabular}
	\caption{\textcolor{black}{Simulation results of the proposed soft gripper grasping a cuboid object and two types of cylindrical objects.}}
	\label{fig:15}
	\vspace{-1em}
\end{figure}

\bibliographystyle{unsrt}
\bibliography{biblio}

\vspace{-3em}
\begin{IEEEbiography}
[{\includegraphics[width=0.9in]{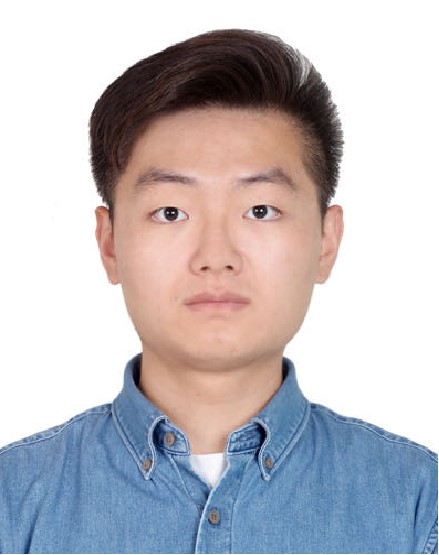}}]
{Xuyang Zhang}
obtained his B.S. degree from Jilin University in 2017 and his M.S. degree from the University of Bristol in 2022. He is currently in his first year of the Ph.D. program in the Department of Engineering at King's College London. His research focuses on the design and simulation of tactile sensors, as well as the design of flexible and rigid grippers integrating tactile sensing.
\end{IEEEbiography}
\vspace{-3em}
\begin{IEEEbiography}
[{\includegraphics[width=0.9in]{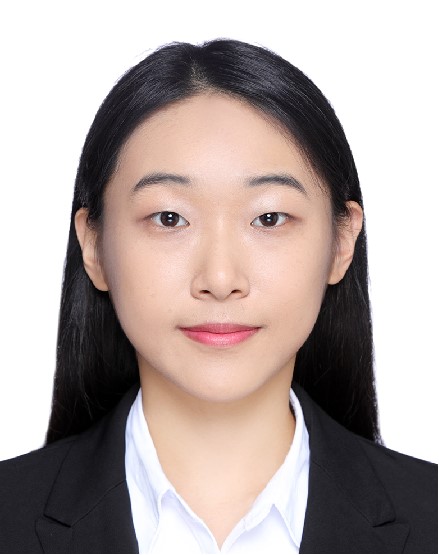}}]
{Tianqi Yang}
received the B.E. degree in electrical engineering from Beijing Jiaotong University, Beijing, China, in 2018, and the M.Sc. degree in electrical and electronic engineering from the University of Bristol, Bristol, U.K., in 2019, where she is currently pursuing the Ph.D. degree with the Visual Information Laboratory. Her current research interests include image analysis and medical imaging.
 
\end{IEEEbiography}
\vspace{-3em}
\begin{IEEEbiography}
[{\includegraphics[width=0.9in]{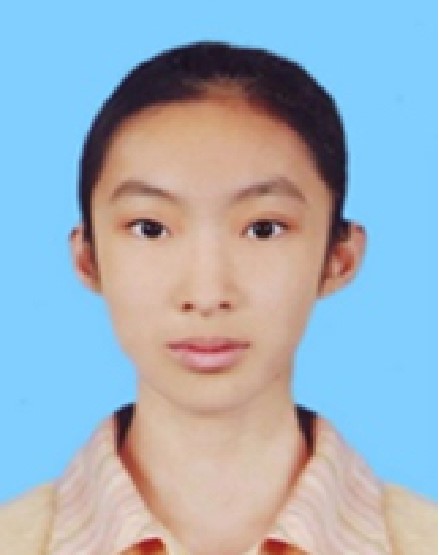}}]
{Dandan Zhang}
is a Lecturer with the Department of Bioengineering at Imperial College London. She also serves as a Lecturer in Artificial Intelligence and Machine Learning as part of the Imperial-X initiative. Dr. Zhang leads the Multi-Scale Embodied Intelligence Lab, where her research focuses on the intersection of robotics, biomedicine, and artificial intelligence. Her lab is committed to developing innovative multi-scale robotic systems with advanced dexterous manipulation capabilities, integrated multi-modality perception. 
\end{IEEEbiography}
\vspace{-3em}
\begin{IEEEbiography}
[{\includegraphics[width=0.9in]{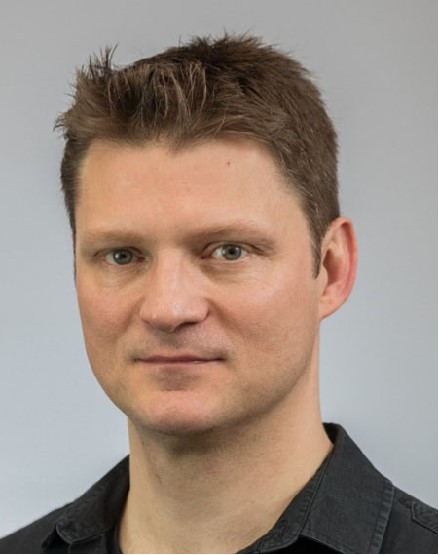}}]
{Nathan Lepora}
received the B.A. degree in Mathematics and the Ph.D. degree in Theoretical Physics from the University of Cambridge, Cambridge, U.K. He is currently a Professor of Robotics and AI with the University of Bristol, Bristol, U.K. He leads the Dexterous Robotics Group in Bristol Robotics Laboratory. Prof. Lepora is a recipient of a Leverhulme Research Leadership Award on ‘A Biomimetic Forebrain for Robot Touch’. He coedited the book “Living Machines” that won the 2019 BMA Medical Book Awards (basic and clinical sciences category). His research team won the ‘University Research Project of the Year’ at the 2022 Elektra Awards.
\end{IEEEbiography}

\end{document}